\documentclass[runningheads]{llncs}
\usepackage{graphicx}
\usepackage{comment}
\usepackage{amsmath,amssymb} % define this before the line numbering.

\DeclareMathOperator*{\argmin}{arg\,min}
\usepackage{color}
\usepackage{multirow}
\usepackage[labelfont=small,bf]{caption}
\usepackage{booktabs}
\usepackage{subcaption}
\usepackage{siunitx}
\usepackage{xcolor}

\newcommand{\bx}{\mathbf{x}}
\newcommand{\bz}{\mathbf{z}}
\renewcommand{\S}{\mathcal{S}}
\newcommand{\Rthree}{\mathbb{R}^3}

\newcommand{\ind}{\mathbbm{1}}
\usepackage{xspace}

\newcommand*{\ie}{i.e.\@\xspace}

\usepackage[normalem]{ulem}

\usepackage{bbm}
\usepackage{tabularx} % in the preamble
\usepackage{hhline}

\usepackage[width=122mm,left=12mm,paperwidth=146mm,height=193mm,top=12mm,paperheight=217mm]{geometry}

\begin{document}
% \linenumbers
\pagestyle{headings}
\mainmatter

\title{Deep Local Shapes: Learning Local SDF Priors for Detailed 3D Reconstruction} % Replace with your title

%  \begin{comment}
% \titlerunning{ECCV-20 submission ID \ECCVSubNumber} 
% \authorrunning{ECCV-20 submission ID \ECCVSubNumber} 
% \author{Anonymous ECCV submission}
% \institute{Paper ID \ECCVSubNumber}
% \end{comment}
%******************

% \begin{comment}
\titlerunning{DeepLS: Learning Local SDF Priors}
% % If the paper title is too long for the running head, you can set
% % an abbreviated paper title here
% % 
\author{Rohan Chabra\inst{1,3, }\thanks{Work performed during an internship at Facebook Reality Labs.} \and
Jan E. Lenssen\inst{2,3, \star} \and
Eddy Ilg \inst{3}\and
Tanner Schmidt \inst{3}\and
Julian Straub \inst{3}\and
Steven Lovegrove \inst{3}\and
Richard Newcombe \inst{3}}
% % 
\authorrunning{Chabra et al.}
\institute{University of North Carolina at Chapel Hill \and
TU Dortmund University \and
Facebook Reality Labs}
% \end{comment}
%******************
\maketitle
\vspace{-1.1cm}
\begin{figure}
\centering
\includegraphics[width=\textwidth]{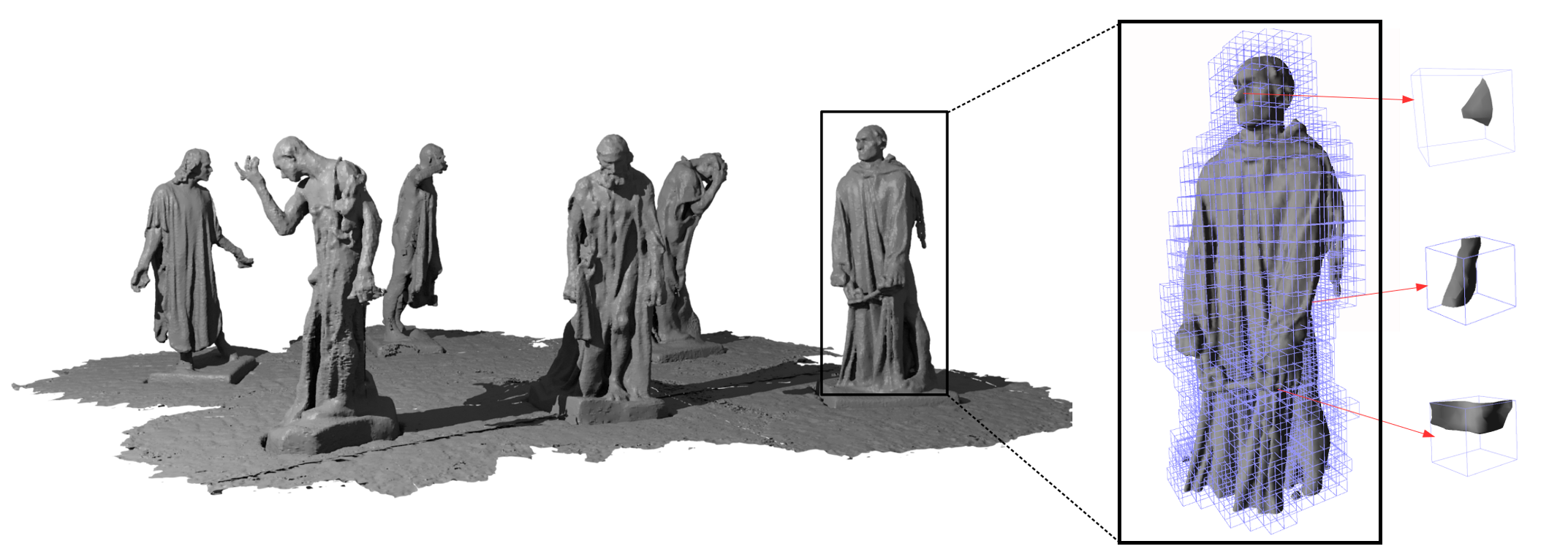}
    \caption{\small Reconstruction performed by our Deep Local Shapes (DeepLS) of the Burghers of Calais scene~\cite{zhou2013dense}. DeepLS represents surface geometry as a sparse set of local latent codes in a voxel grid, as shown on the right. Each code compresses a local volumetric SDF function, which is reconstructed by an implicit neural network decoder.}
    \label{fig:TeaserDeepSDF}
    \vspace{-3mm}
   
\end{figure}
\vspace{-1.1cm}
\begin{abstract}
Efficiently reconstructing complex and intricate surfaces at scale is a long-standing goal in machine perception.
To address this problem we introduce Deep Local Shapes (DeepLS), a deep shape representation that enables encoding and reconstruction of high-quality 3D shapes without prohibitive memory requirements. 
DeepLS replaces the dense volumetric signed distance function (SDF) representation used in traditional surface reconstruction systems with a set of locally learned continuous SDFs defined by a neural network, inspired by recent work such as DeepSDF.
Unlike DeepSDF, which represents an object-level SDF with a neural network and a single latent code, we store a grid of independent latent codes, each responsible for storing information about surfaces in a small local neighborhood.
This decomposition of scenes into local shapes simplifies the prior distribution that the network must learn, and also enables efficient inference.
We demonstrate the effectiveness and generalization power of DeepLS by showing object shape encoding and  reconstructions of full scenes, where DeepLS delivers high compression, accuracy, and local shape completion.
%\keywords{3D reconstruction, local priors, deep implicit surfaces, local DeepSDF}
\end{abstract}

\section{Introduction}
A signed distance function (SDF) represents three-dimensional surfaces as the zero-level set of a continuous scalar field. 
This representation has been used by many classical methods to represent and optimize geometry based on raw sensor observations ~\cite{curless1996volumetric,klein2007parallel,stuhmer2010real,newcombe2010live,newcombe2011kinectfusion}. 
In a typical use case, an SDF is approximated by storing values on a regularly-spaced voxel grid and computing intermediate values using linear interpolation. 
Depth observations can then be used to infer these values and a series of such observations are combined to infer the most likely SDF using a process called fusion.

Voxelized SDFs have been widely adopted and used successfully in a number of applications, but they have some fundamental limitations.
First, the dense voxel representation requires significant amounts of memory (typically on a resource-constrained parallel computing device), which imposes constraints on resolution and the spatial extent that can be represented.
These limits on resolution, as well as sensor limitations, typically lead to surface estimates that are missing thin structures and fine surface details.
Second, as a non-parametric representation, SDF fusion can only infer surfaces that have been directly observed. 
Some surfaces are difficult or impossible for a typical range sensor to capture, and observing every surface in a typical environment is a challenging task. 
As a result, reconstructions produced by SDF fusion are often incomplete.

Recently, deep neural networks have been explored as an alternative representation for signed distance functions.
According to the universal approximation theorem \cite{hornik1989multilayer}, a neural network can be used to approximate any continuous function, including signed distance functions ~\cite{mescheder2019occupancy,park2019deepsdf,michalkiewicz2019deep,Chen2019Implicit}.
With such models, the level of detail that can be represented is limited only by the capacity and architecture of the network.
In addition, a neural network can be made to represent not a single surface but a family of surfaces by, for example, conditioning the function on a latent code.
Such a network can then be used as a parametric model capable of estimating the most likely surface given only partial noisy observations.
Incorporating shape priors in this way allows us to move from the maximum likelihood (ML) estimation of classical reconstruction techniques to potentially more robust reconstruction via maximum a posteriori (MAP) inference.

These neural network representations have their own limitations, however.
Most of the prior work on learning SDFs is object-centric and does not trivially scale to the detail required for scene-level representations. 
This is likely due to the global co-dependence of the SDF values at any two locations in space, which are computed using a shared network and a shared parameterization.
Furthermore, while the ability of these networks to learn distributions over classes of shapes allows for robust completion of novel instances from known classes, it does not easily generalize to novel classes or objects, which would be necessary for applications in scene reconstruction. In scanned real-world scenes, the diversity of objects and object setups is usually too high to be covered by an object-centric training data distribution.

\paragraph{Contribution.} In this work, we introduce Deep Local Shapes (DeepLS) to combine the benefits of both worlds, exposing a trade-off between the prior-based MAP inference of memory efficient deep global representations (e.g., DeepSDF), and the detail preservation of computationally efficient, explicit volumetric SDFs. We divide space into a regular grid of voxels, each with a small latent code representing signed distance functions in local coordinate frames and making use of learned local shape priors. These voxels can be larger than is typical in fusion systems without sacrificing on the level of surface detail that can be represented (c.f. Sec. \ref{sec:icl-nuim}), increasing memory efficiency. The proposed representation has several favorable properties, which are verified in our evaluations on several types of input data:
\begin{enumerate}
    \item It relies on readily available local shape patches as training data and generalizes to a large variety of shapes,
    \item provides significantly finer reconstruction and orders of magnitude faster inference than global, object-centric methods like DeepSDF, and
    \item outperforms existing approaches in dense 3D reconstruction from partial observations, showing thin details with significantly better surface completion and high compression.
\end{enumerate}

\section{Related Work}

The key contribution of this paper is the application of learned local shape priors for reconstruction of 3D surfaces. 
This section will therefore discuss related work on traditional representations for surface reconstruction, learned shape representations, and local shape priors.

\subsection{Traditional Shape Representations}

Traditionally, scene representation methods can broadly be categorized into two categories, namely local and global approaches. \\
\textbf{Local approaches.} Most implicit surface representations from unorganized point sets are based on Blinn's idea of blending local implicit primitives~\cite{blinn1982generalization}. Hope et al.~\cite{hoppe1992surface} explicitly defined implicit surfaces by the tangent of the normals of the input points. Ohtake et al.~\cite{ohtake2005multi} established more control over the local shape functions using quadratic surface fitting and blended these in a multi-scale partition of unity scheme. Curless and Levoy~\cite{curless1996volumetric} introduced volumetric integration of scalar approximations of implicit SDFs in regular grids. This technique was further extended into real-time systems~\cite{klein2007parallel,stuhmer2010real,newcombe2010live,newcombe2011kinectfusion}. Surfaces are also shown to be represented by surfels, i.e. oriented planar surface patches
~\cite{pfister2000surfels,keller2013real,whelan2015elasticfusion}. \\
\textbf{Global approaches.} Global implicit function approximation methods aim to approximate single continuous signed distance functions using, for example, kernel-based techniques ~\cite{carr2001reconstruction,kazhdan2006poisson,ummenhofer2015global,fuhrmann2014floating}. Visibility or free space methods estimate which subset of 3D space is occupied, often by subdividing space into distinct tetrahedra ~\cite{labatut2009robust,jancosek2011multi,aroudj2017visibility}. These methods aim to solve for a globally view consistent surface representation.

Our work falls into the local surface representation category. It is related to the partition of unity approach~\cite{ohtake2005multi}, however, instead of using quadratic functions as local shapes, we use data-driven local priors to approximate implicit SDFs, which are robust to noise and can locally complete supported surfaces. While we also experimented with partition of unity blending of neighboring local shapes, we found it to be not required in practice, since our training formulation already includes border consistency (c.f. Sec \ref{sec:border_consistency}), thus saving function evaluations during decoding. 
In comparison to volumetric SDF integration methods, such as SDF Fusion \cite{newcombe2011kinectfusion}, our approach provides better shape completion and denoising, while at the same time uses less memory to store the representation. Unlike point- or surfel-based methods, our method leads to smooth and connected surfaces.

\subsection{Learned Shape Representations}
Recently there has been lot of work on 3D shape learning using deep neural networks. 
This class of work can also be classified into four categories: point-based methods, mesh-based methods, voxel-based methods and continuous implicit function-based methods.

\textbf{Points.} The methods use generative point cloud models for scene representation ~\cite{achlioptas2017learning,yang2017foldingnet,yuan2018pcn}.
Typically, a neural network is trained to directly regress 3D coordinates of points in the point cloud.

\textbf{Voxels.} These methods provide non-parametric shape representation using 3D voxel grids which store either occupancy~\cite{wu20153d,choy20163d} or SDF information~\cite{dai2017shape,stutz2018learning,liao2018deep}, similarly to the traditional techniques discussed above. 
These methods thus inherit the limitations of traditional voxel representations with respect to high memory requirements. 
Octree-based methods~\cite{tatarchenko2017octree,riegler2017octnet,hane2017hierarchical} relax the compute and memory limitations of dense voxel methods to some degree and have been shown on voxel resolutions of up to $512^3$.

\textbf{Meshes.} These methods use existing~\cite{sinha2016deep} or learned~\cite{groueix2018atlasnet,ben2018multi} parameterization techniques to describe 3D surfaces by morphing 2D planes. 
When using mesh representations, there is a tradeoff between the ability to support arbitrary topology and the ability to reconstruct smooth and connected surfaces. 
Works such as ~\cite{sinha2016deep,ben2018multi} are variations on deforming a sphere into more complex 3D shape, which produces smooth and connected shapes but limits the topology to shapes that are homeomorphic to the sphere. 
AtlasNet, on the other hand, warps multiple 2D planes into 3D which together form a shape of arbitrary topology, but this results in disconnected surfaces. 
Other works, such as Scan2Mesh~\cite{dai2019scan2mesh} and Mesh-RCNN\cite{gkioxari2019mesh}, use deep networks to predict meshes corresponding to range scans or RGB images, respectively.

\textbf{Implicit Functions.} Very recently, there has been significant work on learning continuous implicit functions for shape representations. 
Occupancy Networks~\cite{mescheder2019occupancy} and PiFU \cite{saito2019pifu} represent shapes using continuous indicator functions which specify which subset of 3D space the shapes occupy. 
Similarly, DeepSDF~\cite{park2019deepsdf} approximates shapes using Signed Distance Fields. 
We adopt the DeepSDF model as the backbone architecture for our local shape network. 

Much of the work in this area has focused on learning object-level representations.
This is especially useful when given partial observations of a known class, as the learned priors can often complete the shape with surprising accuracy.
However, this also introduces two key difficulties.
First, the object-level context means that generalization will be limited by the extent of the training set -- objects outside of the training distribution may not be well reconstructed.
Second, object-level methods do not trivially scale to full scenes composed of many objects as well as surfaces (e.g. walls and floors).
In contrast, DeepLS maintains separate representations for small, distinct regions of space, which allows it to scale easily. 
Furthermore, the local representation makes it easier to compile a representative training set; at a small enough scale most surfaces have similar structure.

\subsection{Local Shape Priors}
In early work on using local shape priors, Gal et al.~\cite{gal2007surface} used a database of local surface patches to match partial shape observations. However, the ability to match general observations was limited by the size of the database as the patches could not be interpolated.
Ricao et al.~\cite{ricao2017compressed} used both PCA and a learned autoencoder to map SDF subvolumes to lower-dimensional representations, approaching local shape priors from the perspective of compression. 
With this approach the SDF must be computed by fusion first, which serves as an information bottleneck limiting the ability to develop priors over fine-grained structures.
In another work, Xu et al.~\cite{xu2019disn} developed an object-level learned shape representation using a network that maps from images to SDFs .
This representation is conditioned on and therefore not independent of the observed image.
Williams et al.~\cite{williams2019deep} showed recently that a deep network can be used to fit a representation of a surface by training and evaluating on the same point cloud, using a local chart for each point which is then combined to form a surface atlas.
Their results are on complete point clouds in which the task is simply to densify and denoise, whereas we also show that our priors can locally complete surfaces that were not observed.
Other work on object-level shape representation has explored representations in which shapes are composed of smaller parts. 
Structured implicit functions used anisotropic Gaussian kernels to compose global implicit shape representations \cite{genova2019learning}. Similarly, CvxNets compose shapes using a collection of convex subshapes \cite{deng2019cvxnets}. 
Like ours, both of these methods show the promise of compositional shape modelling, but surface detail was limited by the models used.
Last, concurrent work of Genova et al.~\cite{genova2019learningb} combines a set of irregularly positioned implicit functions to improve details in full object reconstruction. 

\begin{figure*}[t]
  	\centering
  	\includegraphics[width=1.0\textwidth]{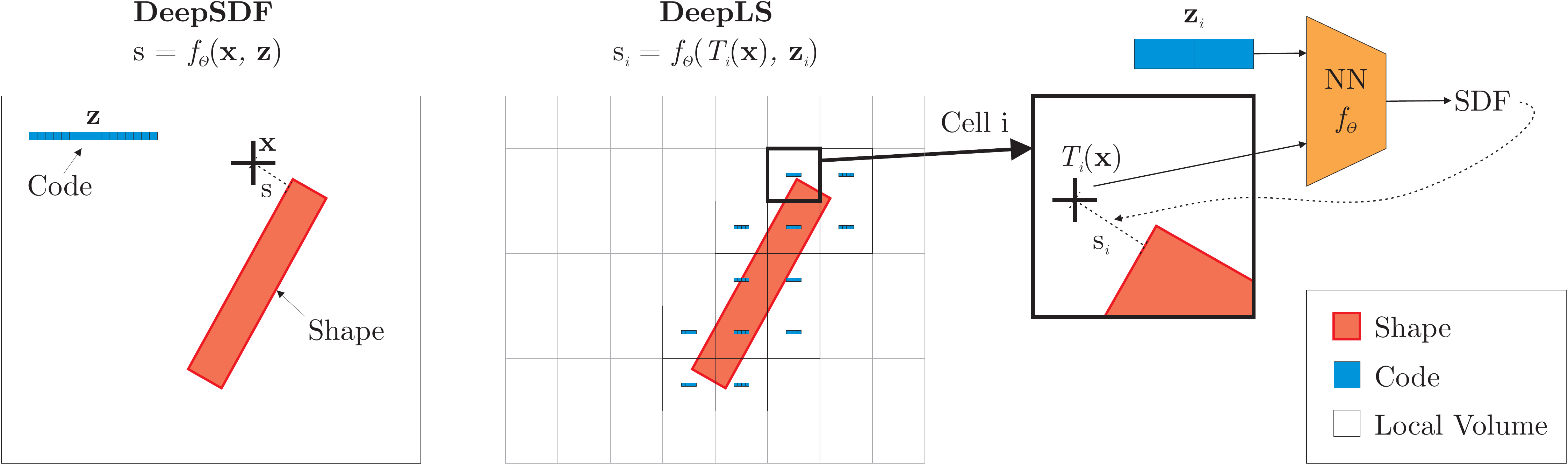}
  	\caption{ \small
        2D example of DeepSDF~\cite{park2019deepsdf} and DeepLS (ours). DeepSDF provides global shape codes (left). We use the DeepSDF idea for local shape codes (center). Our approach requires a matrix of low-dimensional code vectors which in total require less storage than the global version. The gray codes are an indicator for empty space. The SDF to the surface is predicted using a fully-connected network that receives the local code and coordinates as input. 
     	\label{fig:DeepLSArch}
    }
\end{figure*}

\section{Review of DeepSDF}
We will briefly review DeepSDF~\cite{park2019deepsdf}. Let $f_\theta (\bx, \bz)$ be a signed surface distance function modeled as a fully-connected neural network with trainable parameters $\theta$ and shape code $\bz$. Then a shape $\S$ is defined as the zero level set of $f_\theta (\bx, \bz)$:
\begin{equation}
    \S = \{\bx \in \Rthree \mid f_\theta (\bx, \bz) = 0 \} \,.
\end{equation}
In order to simultaneously train for a variety of shapes, a $\bz$ is optimized for each shape while network parameters $\theta$ are shared for the whole set of shapes.

\section{Deep Local Shapes}
The key idea of DeepLS is to compose complex general shapes and scenes from a collection of
simpler local shapes as depicted in Fig.~\ref{fig:DeepLSArch}.
Scenes and shapes
of arbitrary complexity cannot be described with a compact
fixed length shape code such as used by DeepSDF. 
Instead it is more efficient and flexible to encode the space of smaller local
shapes and to compose the global shape from an adaptable amount of local codes.

To describe a surface $\S$ in $\mathbb{R}^3$ using DeepLS, we first
define a partition of the space into local volumes $V_i \subseteq \mathbb{R}^3$
with associated local coordinate systems.
Like in DeepSDF, but at a local level, we describe the surface in each local volume using a code $\bz_i$.
With the transformation $T_i(\bx)$ of the global location $\bx$ into the local
coordinate system, the global surface $\S$ is described as the zero level set
\begin{equation}
  \S = \left\{\bx \in \Rthree \mid \textstyle\bigoplus_i w(\bx, V_i) f_\theta
  \left(T_i(\bx), \bz_i \right) = 0 \right\} \,,
\end{equation}
where $w(\bx, V_i)$ weighs the contribution of
the $i$th local shape to the global shape $\S$, $\bigoplus$ combines the
contributions of local shapes, and $f_{\theta}$ is a shared
autodecoder network for local shapes with trainable parameters $\theta$. 
Various ways of designing the combination operation and weighting function can
be explored. From voxel-based tesselations of the space to more RBF-like
point-based sampling to -- in the limit -- collapsing the volume of a local code
into a point and thus making $\bz_i$ a continuous function of the global space.
\begin{figure}[t]
  	\centering
  	\includegraphics[width=0.7\linewidth]{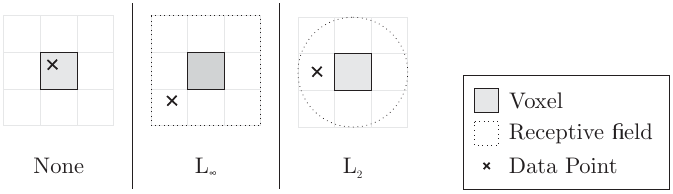}
 	\caption{ \small
        Square ($L_\infty$ norm) and spherical ($L_2$ norm) 
for the extended receptive fields for training local codes.          
    	\label{fig:recp_field}
    }
\end{figure}

Here we focus on exploring the straight forward way of defining local shape codes over
a sparsely allocated voxels $V_i$ of the 3D space as illustrated in Fig.~\ref{fig:DeepLSArch}. 
%Hence,
We define $T_i(\bx) := \bx - \bx_i$, transforming a global point $\bx$ into the local coordinate system of voxel $V_i$ by subtracting its center $\bx_i$. 
The weighting function becomes the indicator function over the volume of voxel
$V_i$.
Thus, DeepLS describes the global surface $\S$ as:
\begin{align}
  \S = \left\{\bx \in \Rthree \mid \textstyle \sum_i \ind_{\bx \in V_i} f_\theta
  \left(T_i(\bx), \bz_i \right) = 0 \right\} \,.
\end{align}

\subsection{Shape Border Consistency}
\label{sec:border_consistency}
We found that with the proposed division of space (i.e. disjoint voxels for local shapes) leads to inconsistent surface estimates at the voxel boundaries.
One possible solution is to choose $w$ as partition of unity \cite{ohtake2005multi} basis functions with local support to combine the decoded SDF values. We experimented with trilinear interpolation as an instance of this. However, this method increases the number of required decoder evaluations to query an SDF value by a factor of eight. 

Instead, we keep the indicator function and train decoder weights and codes such that a local shape is correct beyond the bounds of one voxel, by using training pairs from neighboring voxels.
Then, the SDF values on the voxel boundaries are accurately computable from any of the abutting local shapes. 
We experimented with spheres (\ie $L_2$ norm) and voxels (\ie $L_\infty$ norm) (c.f. Fig.~\ref{fig:recp_field})
for the definition range of extended local shapes and found that using an $L_\infty$ norm with a radius of $1.5$ times the voxel side-length provides a good trade-off between accuracy (fighting border artifacts) and efficiency (c.f. Sec.~\ref{sec:experiments}).

\subsection{Deep Local Shapes Training and Inference}

Given a set of SDF pairs $\{(\bx_j, s_j)\}_{j = 1}^N$, sampled from a set
of training shapes, we aim to
optimize both the parameters $\theta$ of the shared shape decoder
$f_\theta(\cdot)$ and all local shape codes $\{\bz_i\}$ during training and
only the codes during inference.

Let $\mathcal{X}_i = \{\bx_j \mid L(T_i(\bx_j)) < r\}$ denote the set of all training
samples $\bx_j$, falling within a radius $r$ of voxel $i$ with local code $\bz_i$ under the distance metric $L$. 
We train DeepLS by minimizing the negative log
posterior over the training data $\mathcal{X}_i$:
\begin{equation*}
    \argmin_{\theta, \{\bz_i\}} \sum_i \sum_{\bx_j \in \mathcal{X}_i} 
    || f_\theta \left(T_i(\bx_j), \bz_i \right) - s_j ||_1
    + \frac{1}{\sigma^2} ||\bz_i||_2^2  \,.
\end{equation*}
In order to encode a new scene or shape into a set of local codes, we fix
decoder weights $\theta$ and find the maximum a-posteriori codes $\bz_i$ as
\begin{equation}
    \argmin_{\bz_i} \sum_{\bx_j\in \mathcal{X}_i}
        || f_\theta \left(T_i(\bx_j), \bz_i \right) - s_j ||_1 +
    \frac{1}{\sigma^2} ||\bz_i||_2^2 \,,
\end{equation}
given partial observation samples $\{(\bx_j, s_j)\}_{j = 1}^M$ with  $\mathcal{X}_i$ defined as above.

\subsection{Point Sampling}
For sampling data pairs $(\bx_j, s_j)$, we distinguish between sampling from meshes and depth observations. For meshes, the method proposed by Park et al. \cite{park2019deepsdf} is used.
For depth observations, we estimate normals from the depth map and sample points in 3D that are displaced slightly along the normal direction, where the SDF value is assumed to be the magnitude of displacement. In addition to those samples, we obtain free space samples along the observation rays. The process is described formally in the supplemental materials.

\section{Experiments}
\label{sec:experiments}

The experiment section is structured as follows. First, we compare DeepLS against recent deep learning methods (e.g. DeepSDF, AtlasNet) in Sec. \ref{sec:exp_shapenet}. Then, we present results for scene reconstruction and compare them against related approaches on both synthetic and real scenes in Sec. \ref{sec:real_scene_experiments}.

\paragraph{Experiment setup.} The models used in the following experiments were trained on a set of local shape patches, obtained from 200 primitive shapes (e.g. cuboids and ellipsoids) and a total of 1000 shapes from the 3D Warehouse~\cite{Warehouse3D} dataset (200 each for the airplane, table, chair, lamp, and sofa classes). Our decoder is a four layer MLP, mapping from latent codes of size $128$ to the SDF value.
We present examples from the training set, several additional results, comparisons and further details about the experimental setup in the supplemental materials.

\begin{figure}[t]
    \centering 
    \includegraphics[width=1\textwidth]{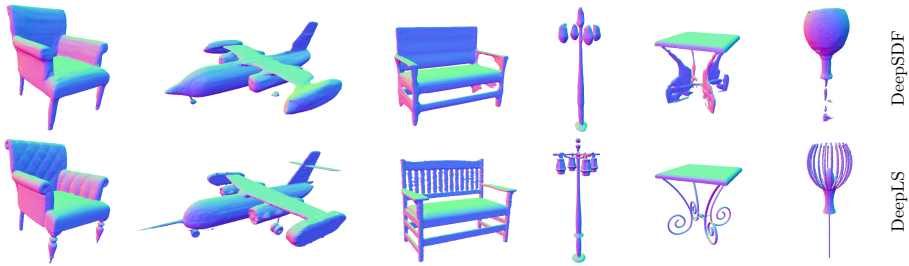}
    \caption{\small Qualitative comparison of DeepLS with DeepSDF on some shapes from the 3D Warehouse~\cite{Warehouse3D} dataset.\label{fig:ShapenetAccuracy}}
\end{figure}

 \begin{table}[t]
\footnotesize
\centering
 \begin{tabular}{l|ccccc|c|c}
 \toprule
  \multicolumn{1}{c|}{\multirow{2}{*}{Method}}& \multicolumn{5}{c|}{\multirow{1}{*}{Shaoe Category}}&
  \multicolumn{1}{c|}{\multirow{1}{*}{Decoder}}& \multicolumn{1}{c}{\multirow{1}{*}{Represent.}}
 \\
 \multicolumn{1}{c|}{} & \multicolumn{1}{c}{chair} &
 \multicolumn{1}{c}{plane}& \multicolumn{1}{c}{table}&
 \multicolumn{1}{c}{lamp}& \multicolumn{1}{c|}{sofa} & \multicolumn{1}{c|}{Params} & \multicolumn{1}{c}{Params} \\\midrule
 AtlasNet-Sph.~\cite{groueix2018atlasnet} & 0.752 & 0.188 & 0.725 & 2.381 & 0.445 & 3.6 M & 1.0 K\\
 AtlasNet-25~\cite{groueix2018atlasnet}    & 0.368 & 0.216 & 0.328 & 1.182 & 0.411& 43.5 M & 1.0 K\\
 DeepSDF~\cite{park2019deepsdf}      &  0.204 & 0.143 & 0.553 & 0.832 & 0.132& 1.8 M & \bf{0.3} K\\
  DeepLS      & \,\, \bf{0.030} \,\,& \,\,\bf{0.018}\,\, & \,\,\bf{0.032} \,\, & \,\,\bf{0.078}\,\, & \,\,\bf{0.044} \,\,& \,\,\bf{0.05 M}\,\, & 312 K \\
 \bottomrule

   \end{tabular}
   \vspace{2mm}
   \caption{\small Comparison for reconstructing shapes from the 3D Warehouse~\cite{Warehouse3D} test set, using the Chamfer distance. Results with additional metrics are similar as detailed in the supplemental materials. Note that due to the much smaller decoder, DeepLS is also orders of magnitudes faster in decoding (querying SDF values). } 
   \label{tab:shapenetTest}
   \vspace{-4mm}
\end{table}

\subsection{Object Reconstruction}
\label{sec:exp_shapenet}

\subsubsection{3D Warehouse~\cite{Warehouse3D}}
We quantitatively evaluate surface reconstruction accuracy of DeepLS and other shape learning methods on various classes from the 3D Warehouse dataset.
Quantitative results for the chamfer distance error are shown in Table~\ref{tab:shapenetTest}.
As can be seen DeepLS improves over related approaches by approximately one
order of magnitude. It should be noted that this is not a comparison between equal methods since the other methods infer a global, object-level representation that comes with other advantages. Also, the parameter distribution varies significantly (c.f. Tab.~ \ref{tab:shapenetTest}).  Nonetheless, it proves that local shapes lead to superior reconstruction quality and that implicit functions modeled by a deep neural network are capable of representing fine details.
Qualitatively, DeepLS encodes and
reconstructs much finer surface details as can be seen in Fig.~\ref{fig:ShapenetAccuracy}.

\begin{figure}
\centering
    \includegraphics[width=0.9\linewidth]{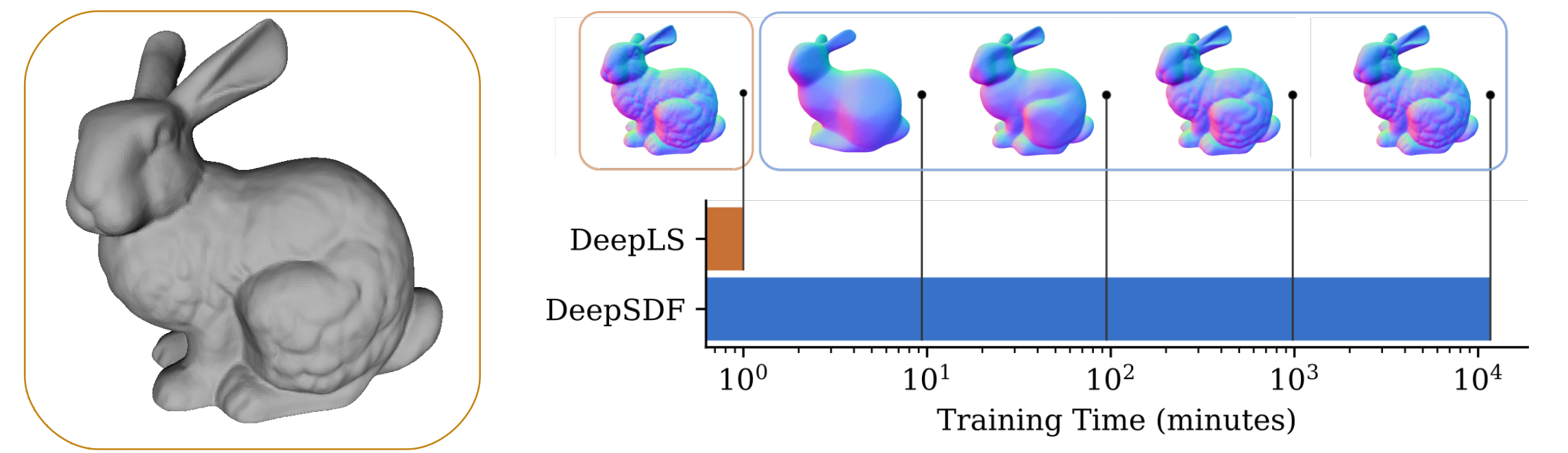}
    \caption{\small A comparison of the efficiency of DeepLS and DeepSDF. With DeepLS, a model trained for one minute is capable of reconstructing the Stanford Bunny~\cite{StanfordScanRep} in full detail. We then trained a DeepSDF model to represent the same signed distance function corresponding to the Stanford Bunny until it reaches the same accuracy. This took over 8 days of GPU time (note the log scale of the plot). \label{fig:DeepLS_vs_DeepSDF}}
    \vspace{-1mm}
\end{figure}
\subsubsection {Efficiency Evaluation on Stanford Bunny~\cite{StanfordScanRep}}
Further, we show the superior inference efficiency of DeepLS with a simple experiment, illustrated in Figure \ref{fig:DeepLS_vs_DeepSDF}.
A DeepLS model was trained on a dataset composed only of randomly oriented primitive shapes. It is used to infer local codes that pose an implicit representation of the Stanford Bunny. Training and inference together took just one minute on a single GPU. The result is an RMSE of only 0.03\% relative to the length of the diagonal of the minimal ground truth bounding box, highlighting the ability of DeepLS to generalize to novel shapes. 
For comparison, we also trained a DeepSDF model to represent only the Stanford Bunny (jointly training latent code and decoder).
In order to achieve the same surface error, this model required over 8 days of GPU time, showing that the high compression rates and object-level completion capabilities of DeepSDF and related techniques comes at the cost of long training and inference times.
This is likely caused at least in part by gradient computation amongst all training samples, which we avoid by subdividing physical space and optimizing local representations in parallel.

\vspace{-3mm}
\subsection{Scene Reconstruction}

We evaluate the ability of DeepLS to reconstruct at scene scale using synthetic (in order to provide quantitative comparisons) and real
depth scans. For synthetic scans, we use the ICL-NUIM RGBD benchmark
dataset~\cite{handa:etal:ICRA2014}. The evaluation on real scans is done using the 3D Scene Dataset \cite{zhou2013dense}. For quantitative evaluation, the asymmetric Chamfer distance metric provided by the benchmark~\cite{handa:etal:ICRA2014} is used.

 \begin{table}[t]
\centering
\footnotesize
 \begin{tabular}{l|c|cccc}
\toprule
  \multicolumn{1}{l|}{Method} & \multicolumn{1}{c|}{mean} & \multicolumn{1}{c}{kt0} & \multicolumn{1}{c}{kt1} & \multicolumn{1}{c}{kt2} & \multicolumn{1}{c}{kt3}  \\
\midrule
 TSDF Fusion   & 5.42 mm &  5.35 mm & 5.88 mm & 5.17 mm & 5.27 mm \\
 DeepLS        &\,\, \bf{4.92 mm} \,\, &  \,\, \bf{5.15 mm}\,\, & \,\,\bf{5.48 mm}\,\, & \,\,\bf{4.32 mm}\,\, & \,\,\bf{4.71 mm}\,\, \\
\bottomrule
   \end{tabular}
   \vspace{2mm}
   \caption{\small Surface reconstruction accuracy of DeepLS and TSDF Fusion~\cite{curless1996volumetric} on the synthetic ICL-NUIM dataset ~\cite{handa:etal:ICRA2014} benchmark} 
   \vspace{-6mm}
   \label{tab:ICLTestTable}
\end{table}

\begin{figure}[t]
\centering
    \begin{subfigure}{0.45\textwidth}
        \centering 
        \includegraphics[width=\textwidth]{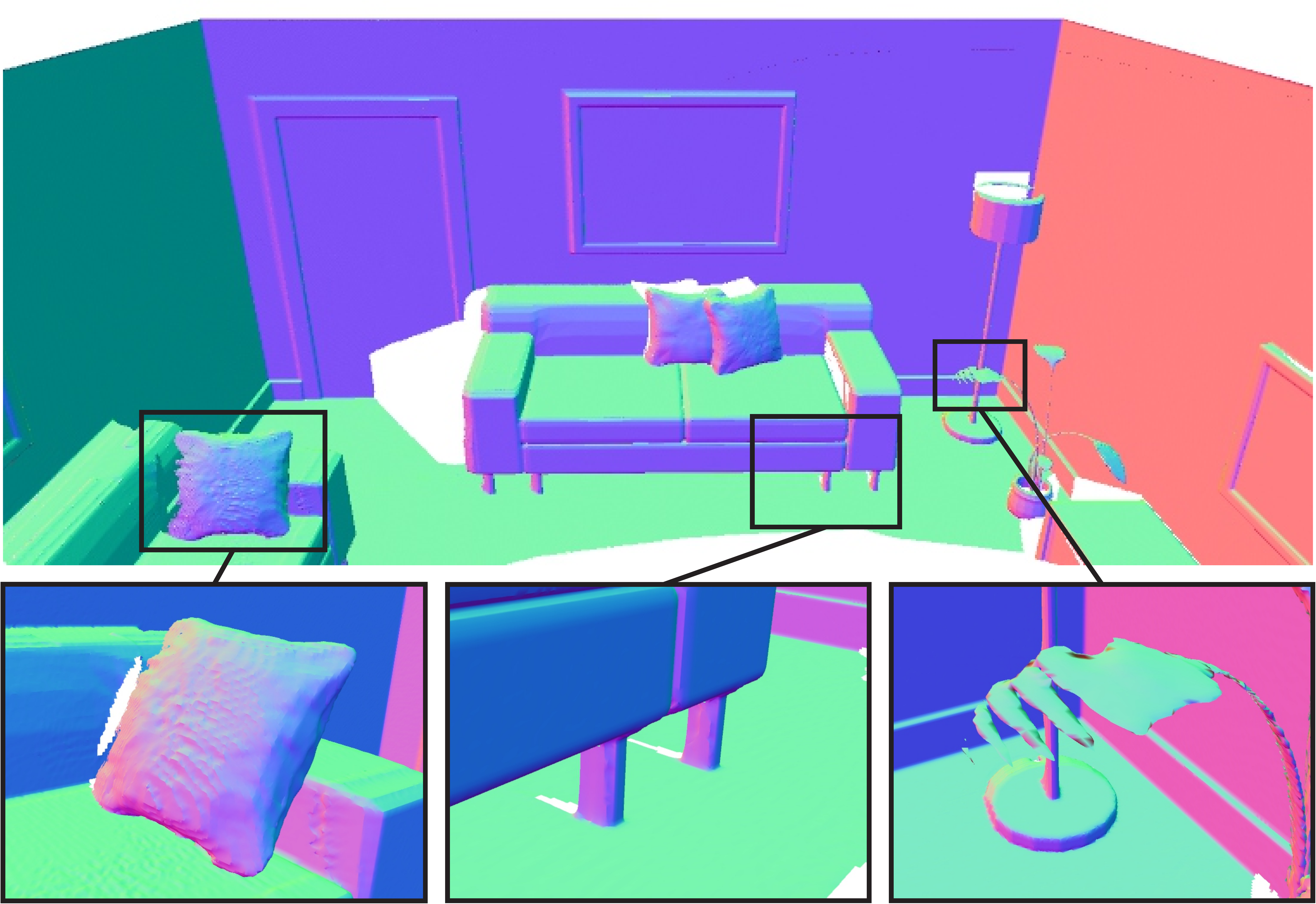}
        \vspace*{-5mm}
        \caption{TSDF Fusion~\cite{newcombe2011kinectfusion}}
    \end{subfigure}
    \hspace{0.5cm}
    \begin{subfigure}{0.45\textwidth}
        \centering 
        \includegraphics[width=\textwidth]{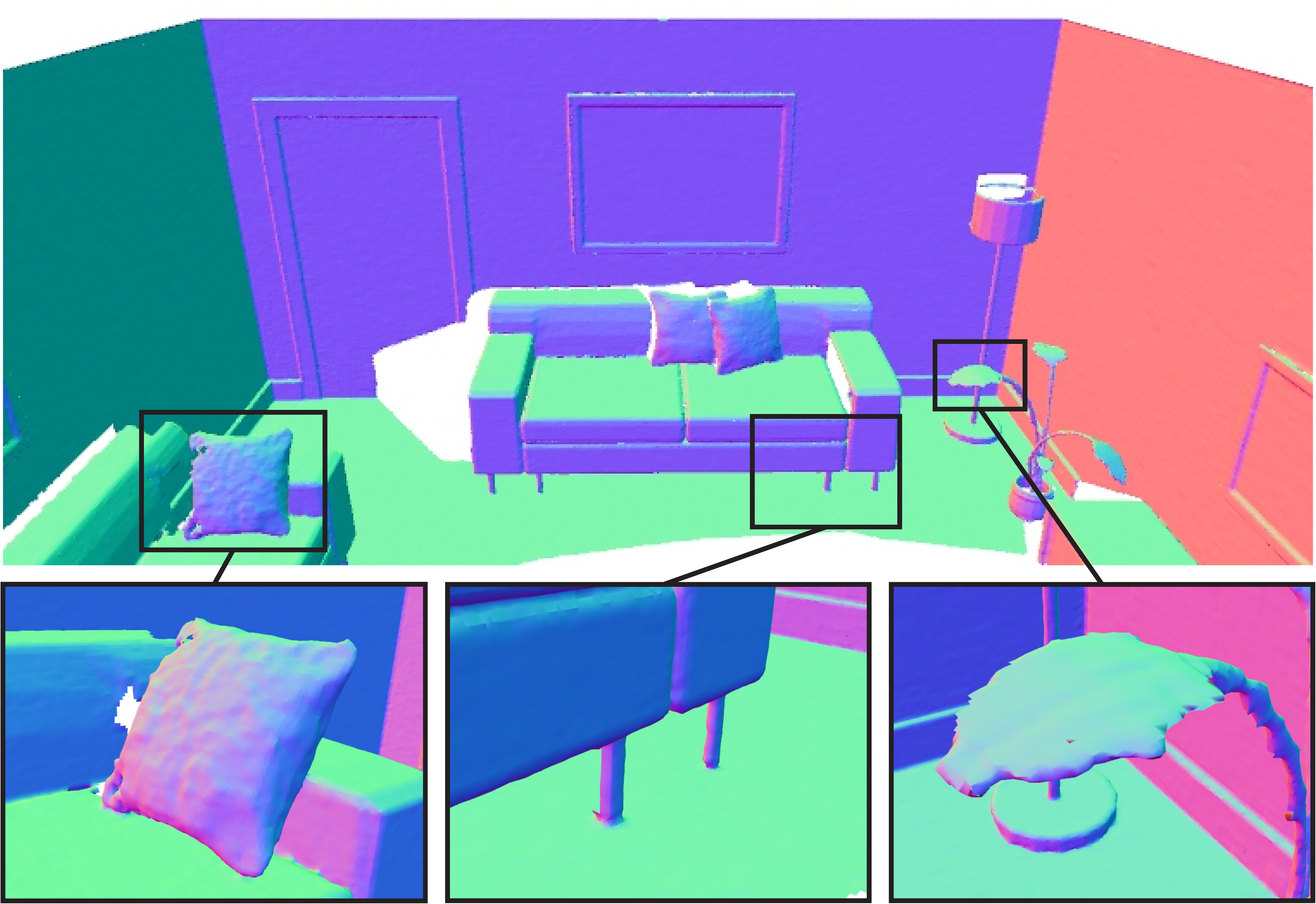}
        \vspace*{-5mm}
        \caption{DeepLS (ours)}
    \end{subfigure}
    
   \vspace{-2mm}
    \caption{\small Qualitative results of TSDF Fusion~\cite{newcombe2011kinectfusion} (left) and DeepLS (right) for scene reconstruction on a synthetic ICL-NUIM~\cite{handa:etal:ICRA2014} (CC BY 3.0, Handa, A., Whelan, T., McDonald, J., Davison) scene. The highlighted areas indicate the ability of DeepLS to handle oblique viewing angles, partial observation, and thin structures.}
    \label{fig:SynthQualResults}
   \vspace{-1mm}
   
\end{figure}

\begin{figure*}[t]
  \centering
   \hspace{-5mm}
\begin{subfigure}[t]{0.32\linewidth}
 \centering
  \includegraphics[width=1.05\textwidth]{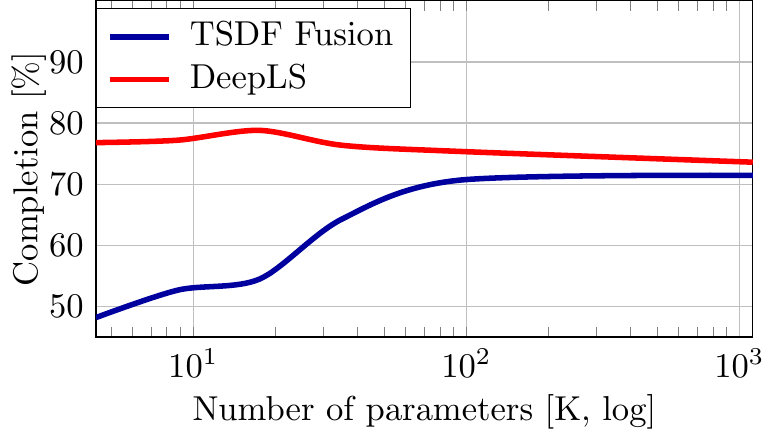}
        \caption{Completion\label{fig:completionICLPlot}}
  \end{subfigure}
\begin{subfigure}[t]{0.32\linewidth}
 \centering
  \includegraphics[width=1.05\textwidth]{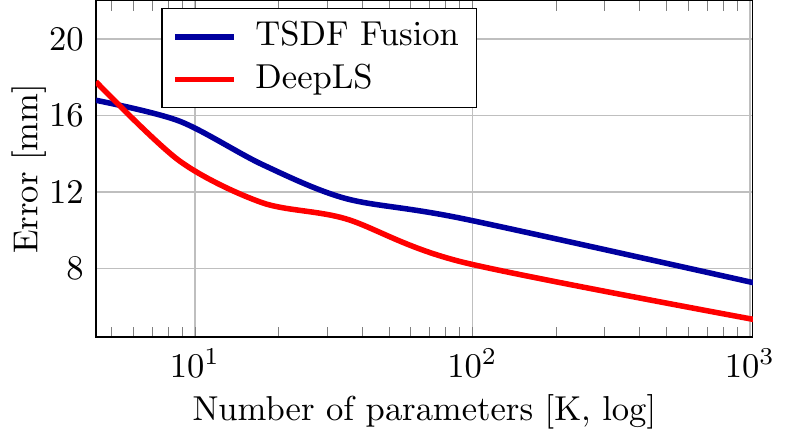}
        \caption{Surface error}
  \end{subfigure}
\begin{subfigure}[t]{0.32\linewidth}
 \centering
    \includegraphics[width=1.02\textwidth]{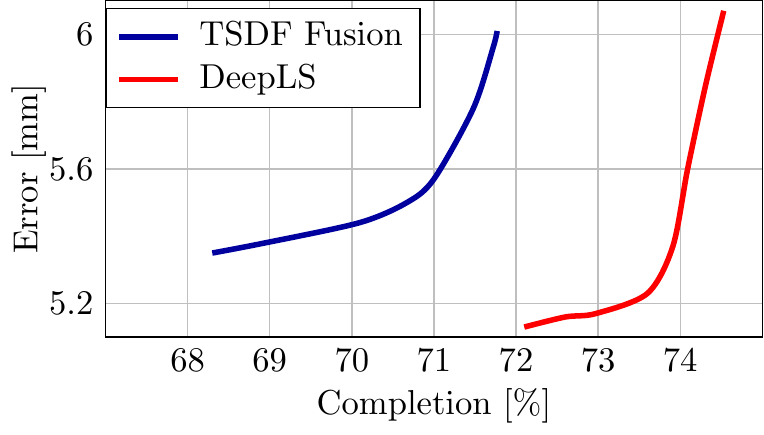}
        \caption{Completion vs. error \label{fig:completion_vs_acc}}
  \end{subfigure}

   \vspace{-1mm}
    \caption{\small Comparison of completion (a) and surface error (b) as a function of representation parameters on a synthetic scene from the ICL-NUIM~\cite{handa:etal:ICRA2014} dataset. 
    In contrast to TSDF Fusion, DeepLS maintains reconstruction completeness almost independent of compression rate. On the reconstructed surfaces
    (which is 50\% less for TSDF Fusion) the surface error decreases for both methods (c.f. Fig.~\ref{fig:Compression}). Plot (c) shows the trend of surface error vs. mesh completion. DeepLS consistently shows higher completion at the same surface error. It scores less error than TSDF Fusion in all but the highest compression setting but it produces nearly two times more complete reconstruction than TSDF Fusion at this compression rate.
   % Both methods have tunable parameters that trade-off accuracy over completion.
    }
   \vspace{-5mm}
    \label{fig:compression_accuracy_plots}
\end{figure*}

\begin{figure}[t]
    \centering 
    \includegraphics[width=1.0\textwidth]{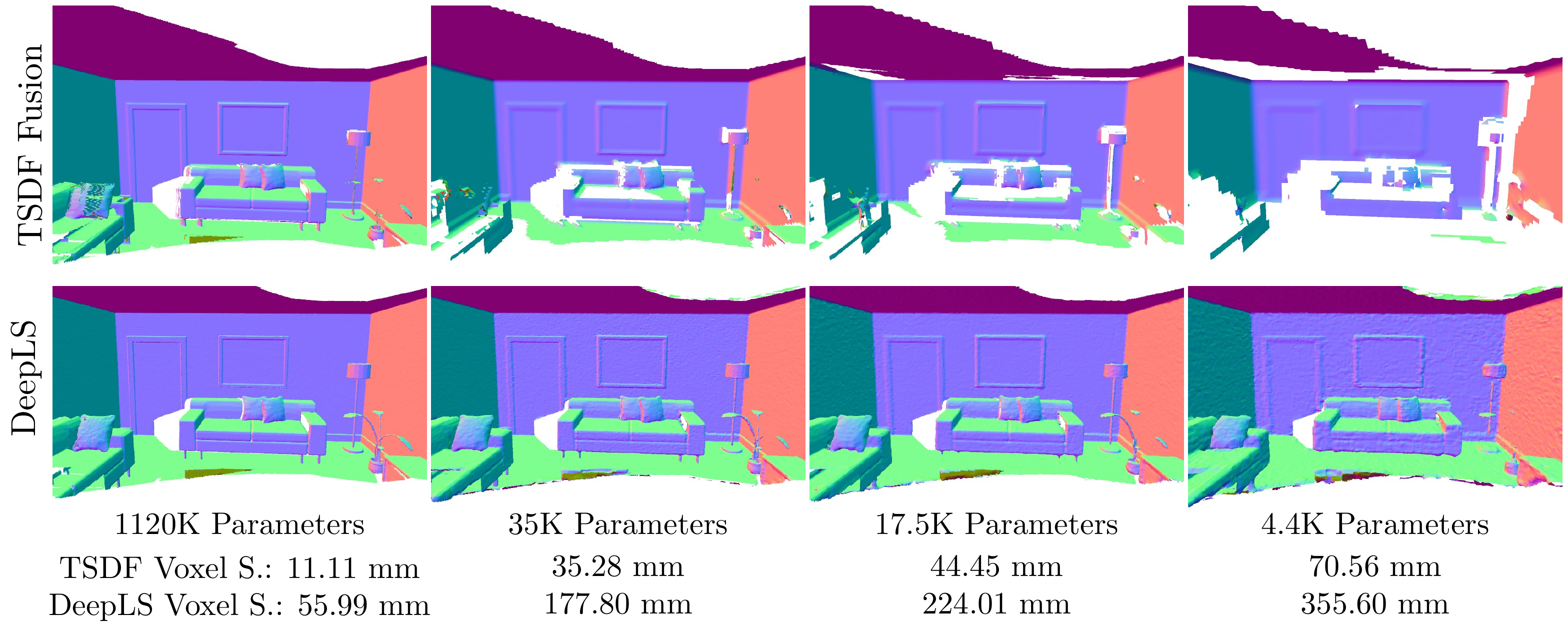}
    \vspace{-7mm}
    \caption{\small Qualitative analysis of representation size with DeepLS and
    TSDF Fusion~\cite{curless1996volumetric} on a synthetic scene in the
    ICL-NUIM~\cite{handa:etal:ICRA2014} dataset. DeepLS is able to retain
    details at higher compression rates (lower number of parameters). 
    It achieves these compression rates by using bigger local shape voxels, leading to a stronger influence of the priors.\label{fig:Compression}}
    \vspace{-2mm}
\end{figure}

\vspace{-0.3cm}
\subsubsection{Synthetic ICL-NUIM Dataset Evaluation\label{sec:icl-nuim}}

We provide quantitative measurements of surface reconstruction quality on all four ICL-NUIM sequences in Table~\ref{tab:ICLTestTable}, where each system has been tuned for lowest surface error. 
We also show results qualitatively in Fig. \ref{fig:SynthQualResults} and show additional results, e.g. on data with artificial noise, in the supplemental materials.
Most surface reconstruction techniques involve a tradeoff between surface accuracy and completeness.
For TSDF fusion systems such as KinectFusion~\cite{newcombe2011kinectfusion}, this tradeoff is driven by choosing a truncation distance and the minimum confidence at which surfaces are extracted by marching cubes.
With DeepLS, we only extract surfaces up to some fixed distance from the nearest observed depth point, and this threshold is what trades off accuracy and completion of our system.
For a full and fair comparison, we derived a pareto-optimal curve by varying these parameters for the two methods on the `kt0` sequence of the ICL-NUIM benchmark and plot the results in  Figure~\ref{fig:compression_accuracy_plots}. 
We measure completion by computing the fraction of ground truth points for which there is a reconstructed point within \SI{7}{\milli\meter}.
Generally, DeepLS can reconstruct more complete surfaces at the same level of accuracy as SDF Fusion.

\begin{figure}[t]
    \centering
    \begin{subfigure}{0.95\textwidth}
    \includegraphics[width=\textwidth]{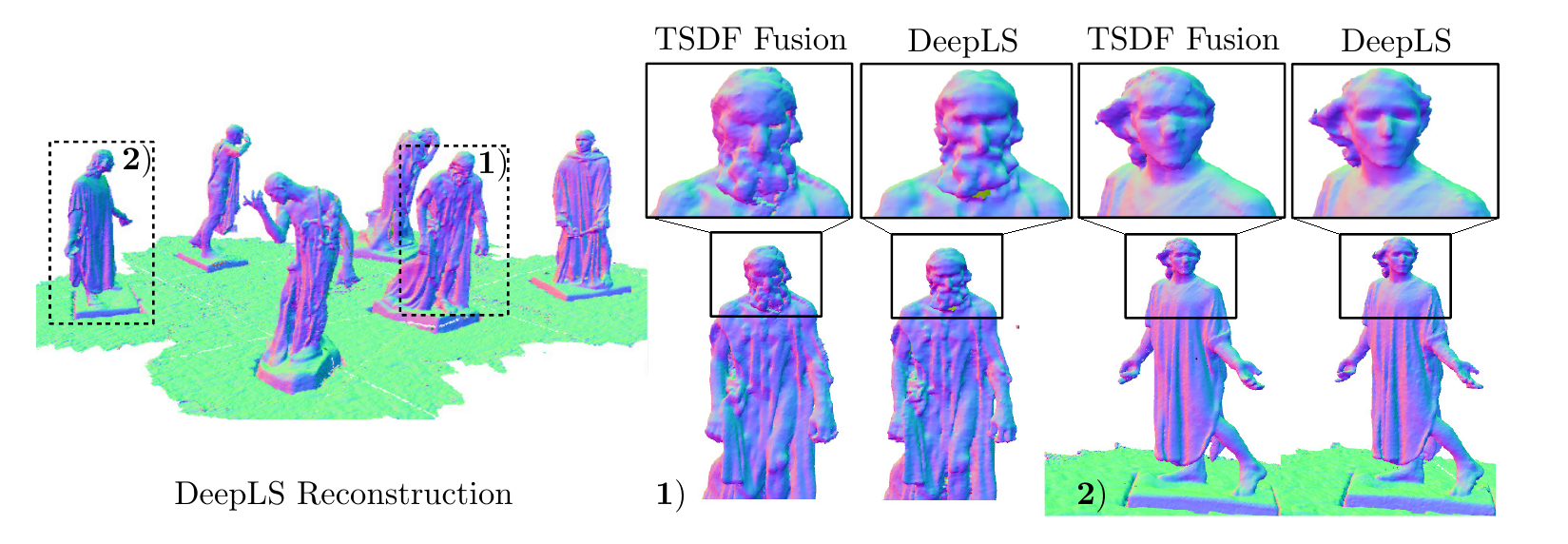}
        \caption{Burghers of Calais scene} 
    \end{subfigure}
    \begin{subfigure}{0.95\textwidth}
    \includegraphics[width=\textwidth]{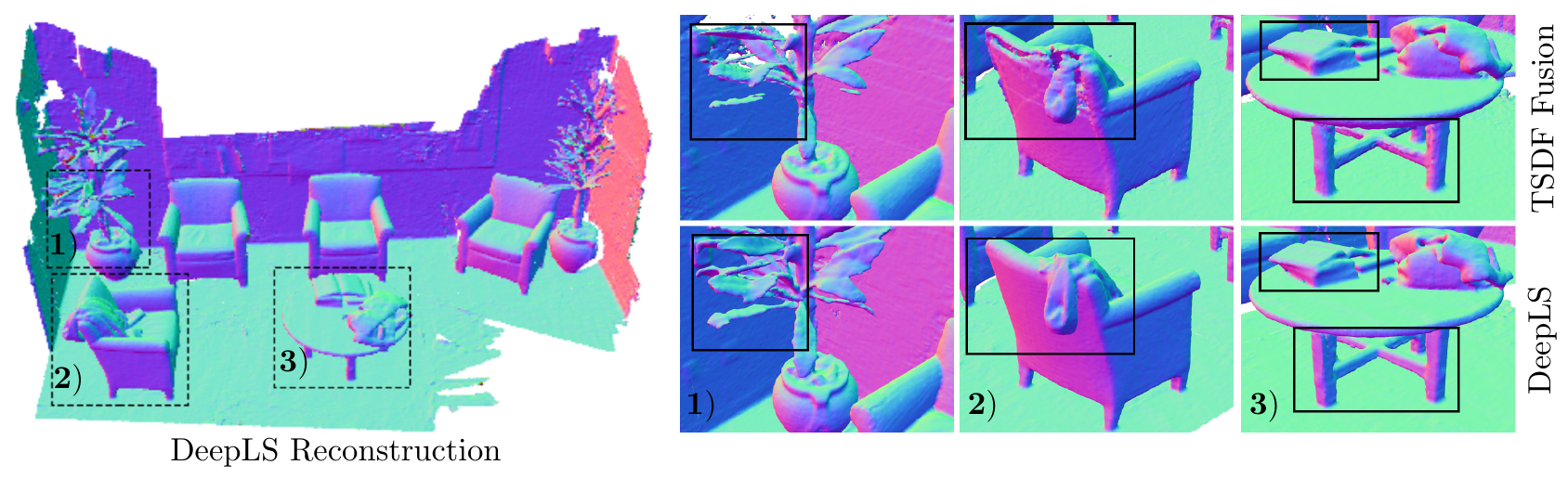}
        \caption{Lounge scene} 
    \end{subfigure}
    \caption{\small Qualitative results for DeepLS and TSDF Fusion~\cite{curless1996volumetric} on two scenes of the 3D Scene Dataset~\cite{zhou2013dense}. Best viewed with zoom in the digital version of the paper.}
   \vspace{-5mm}
    \label{fig:3DScene}
\end{figure}

The number of representation parameters used by DeepLS is theoretically independent of the rendering resolution and only depends on the resolution of the local shapes. In contrast, traditional volumetric scene reconstruction methods such as TSDF Fusion have a tight coupling between number of parameters and the desired rendering resolution.
We investigate the relationship between representation size per unit volume of DeepLS and TSDF Fusion by evaluating the surface error and completeness as a function of the number of parameters. 
As a starting point we choose a representation that uses $8^3$ parameters per $5.6 \textrm{cm} \times 5.6\textrm{cm} \times 5.6\textrm{cm}$  volume ($7$ mm voxel resolution). To increase compression we increase the voxel size for TSDF Fusion and the local shape code volume size for DeepLS.
We provide the quantitative and qualitative analysis of the scene reconstruction results with varying representation size in Fig.~\ref{fig:compression_accuracy_plots}~(a and b) and Fig.~\ref{fig:Compression} respectively.
%Therefore, we scale voxel resolution for KinectFusion and local shape resolution for DeepLS in this study.
The plots in Fig.~\ref{fig:compression_accuracy_plots} show conclusively that TSDF Fusion drops to about 50\% less complete reconstructions while DeepLS maintains completeness even at the highest compression rate, using only $4.4$K parameters for the full scene. 
Quantitatively, TSDF Fusion also achieves low surface error for high compression. However, this can be contributed to the used ICL-NUIM benchmark metric, which does not strongly punish missing surfaces.

\begin{table}[t]
    	\centering
    	\small
    	\footnotesize \setlength\tabcolsep{2.4pt} 
    		\begin{tabular}{c|cc|cc|cc|cc|cc}
    			\toprule
    			\multicolumn{1}{c|}{\multirow{2}{*}{Method}} & \multicolumn{2}{c|}{\multirow{1}{*}{Burghers}} & \multicolumn{2}{c|}{\multirow{1}{*}{Lounge}} &  \multicolumn{2}{c|}{\multirow{1}{*}{CopyRoom}}  &
    			\multicolumn{2}{c|}{\multirow{1}{*}{StoneWall}}  &
    			\multicolumn{2}{c}{\multirow{1}{*}{TotemPole}}   \\
    			%\cline{2-11}
    			
    			\multicolumn{1}{c|}{} & \multicolumn{1}{c}{Error} & \multicolumn{1}{c|}{Comp} & \multicolumn{1}{c}{Error} & \multicolumn{1}{c|}{Comp} & \multicolumn{1}{c}{Error} & \multicolumn{1}{c|}{Comp} & \multicolumn{1}{c}{Error} &
    			\multicolumn{1}{c|}{Comp} & \multicolumn{1}{c}{Error} &
    			\multicolumn{1}{c}{Comp} \\
    			\midrule
    			
    			TSDF F.~\cite{curless1996volumetric} & 10.11 & 85.46 & 11.71 & 85.17 & 12.35 & 83.99 & 14.23 & 91.02 & 13.03 & 83.73 \\
    			%\cline{1-11}
    			DeepLS & \textbf{5.74} & \textbf{95.78} & \textbf{7.38} & \textbf{96.00} & \textbf{10.09} & \textbf{99.70} & \textbf{6.45} & \textbf{91.37} & \textbf{8.97} & \textbf{87.23} \\
    			\bottomrule
    			
    		\end{tabular}
    		\vspace{2mm}
    	\caption{\small Quantitative evaluation of DeepLS with TSDF Fusion on 3D Scene Dataset~\cite{zhou2013dense}. The error is measured in mm and \emph{Comp} (completion) corresponds to the percentage of ground truth surfaces that have reconstructed surfaces within $7$ mm. Results suggest that DeepLS produces more accurate and complete 3D reconstruction in comparison to volumetric fusion methods on real depth acquisition datasets.}
    	\vspace{-7mm}
    	\label{Table:3DScene}
    \end{table}

\vspace{-0.3cm}
\subsubsection{Evaluation on Real Scans}
\label{sec:real_scene_experiments}We evaluate DeepLS on the 3D Scene Dataset~\cite{zhou2013dense}, which contains several scenes captured by commodity structured light sensors, and a challenging scan of thin objects.
In order to also provide quantitative errors we assume the reconstruction performed by volumetric fusion~\cite{curless1996volumetric} of all depth frames to be the \emph{ground truth}. We then apply DeepLS and TSDF fusion on a small subset of depth frames, taking every 10th frame in the capture sequence. The quantitative results of this comparison are detailed in Table~\ref{Table:3DScene} for various scenes. It is shown that DeepLS produces both more accurate and more complete 3D reconstructions. Furthermore, we provide qualitative examples of this experiment in Fig.~\ref{fig:3DScene} for the outdoor scene ``Burghers of Calais" and for the indoor scene ``Lounge". Notice, that DeepLS preserves more details on the faces of the statues in ``Burghers of Calais" scene and reconstructs thin details such as leaves of the plants in ``Lounge" scene. \\

\captionsetup[sub]{font=scriptsize,labelfont={bf,sf}}

\begin{figure}[t]
    \centering 
    \begin{subfigure}{0.22\textwidth}
        \centering 
        \includegraphics[width=\textwidth,clip,trim=10 0 0 0]{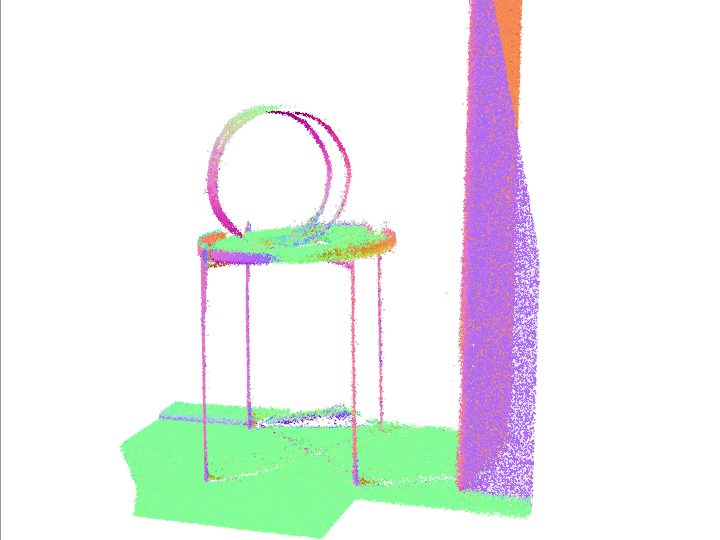}
        \caption{Input Points} 
    \end{subfigure}
    \hfill
    \begin{subfigure}{0.22\textwidth}
        \centering 
        \includegraphics[width=\textwidth]{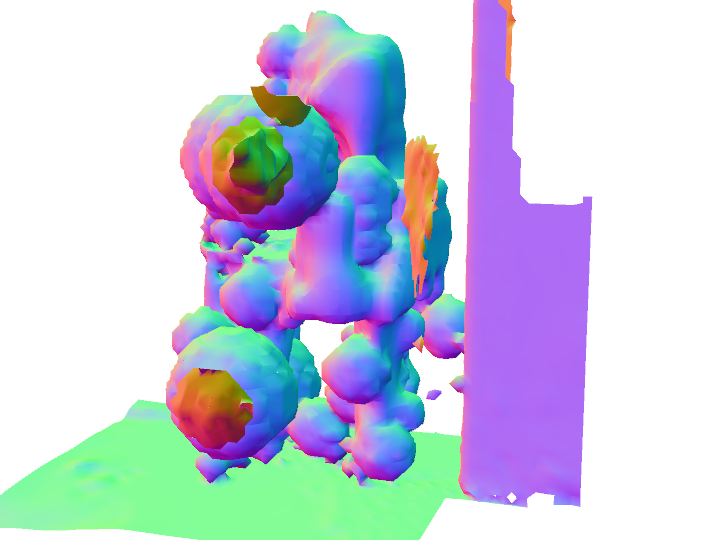}
        \caption{MPU~\cite{ohtake2005multi}}
        \label{subfig:Stool_MPU}
    \end{subfigure}   
    \hfill
    \begin{subfigure}{0.22\textwidth}
        \centering 
        \includegraphics[width=\textwidth]{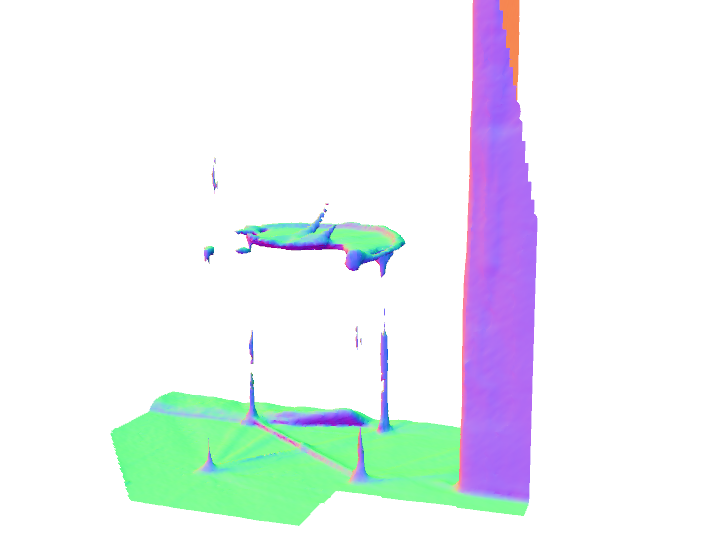}
        \caption{TSDF Fusion~\cite{newcombe2011kinectfusion}}
    \end{subfigure}
    \hfill
    \begin{subfigure}{0.22\textwidth}
        \centering 
        \includegraphics[width=\textwidth]{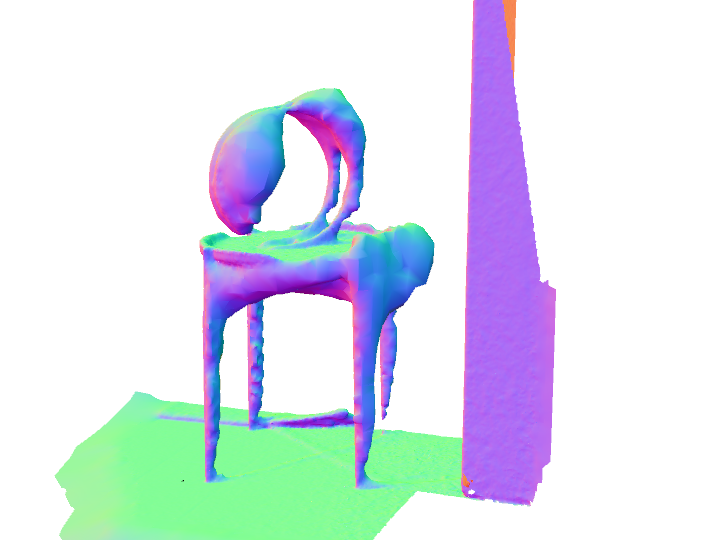}
        \caption{SSD~\cite{calakli2011ssd}}
    \end{subfigure}    
    \begin{subfigure}{0.22\textwidth}
        \centering 
        \includegraphics[width=\textwidth,clip,trim=10 0 0 0]{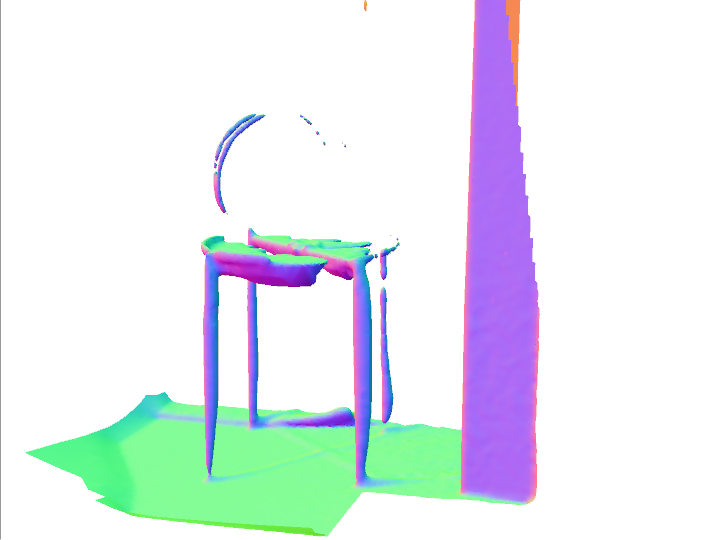}
        \caption{PSR~\cite{kazhdan2013screened}}
    \end{subfigure}
    \hfill
    \begin{subfigure}{0.22\textwidth}
        \centering 
        \includegraphics[width=\textwidth]{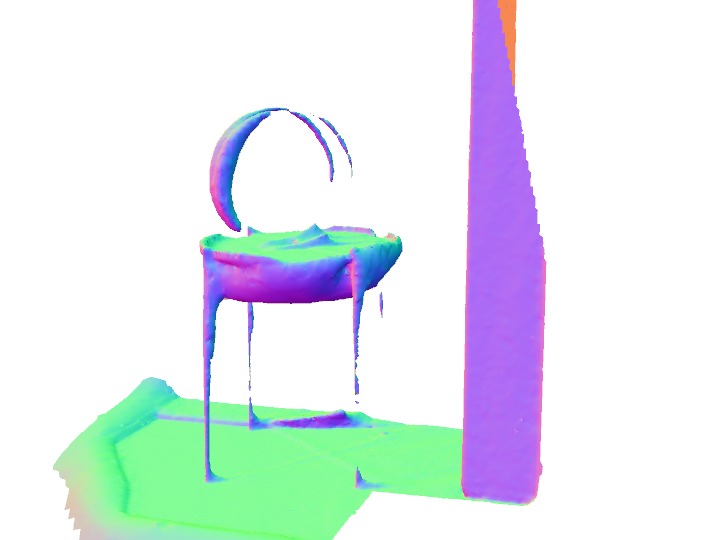}
        \caption{PFS~\cite{ummenhofer2015global}}
    \end{subfigure}
    \hfill
    \begin{subfigure}{0.22\textwidth}
        \centering 
        \includegraphics[width=\textwidth]{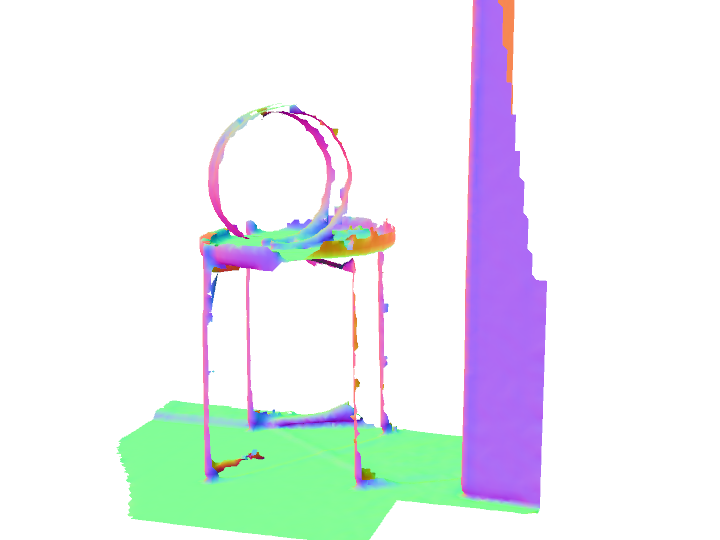}
        \caption{TSR~\cite{aroudj2017visibility}}
    \end{subfigure}
    \hfill
    \begin{subfigure}{0.22\textwidth}
        \centering 
        \includegraphics[width=\textwidth]{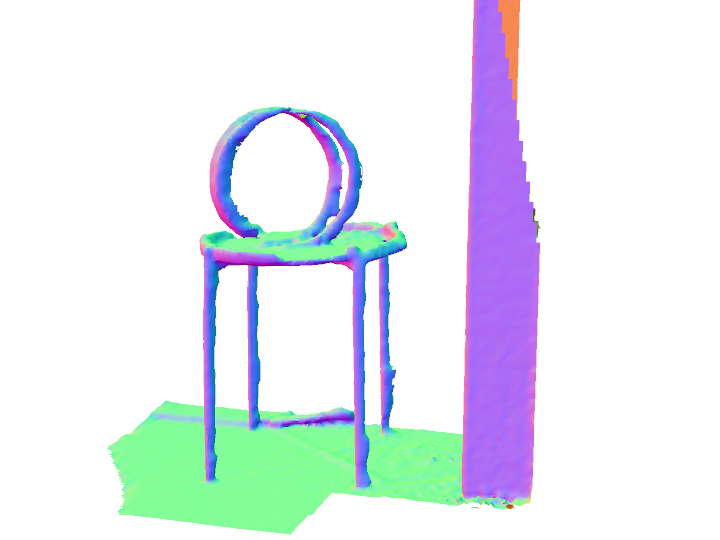}
        \caption{\textbf{DeepLS (Ours)}}
    \end{subfigure}\\[5ex]
   
        \vspace{-5mm}
    
    \begin{subfigure}{0.9\textwidth}
        \centering 
        \includegraphics[width=\textwidth]{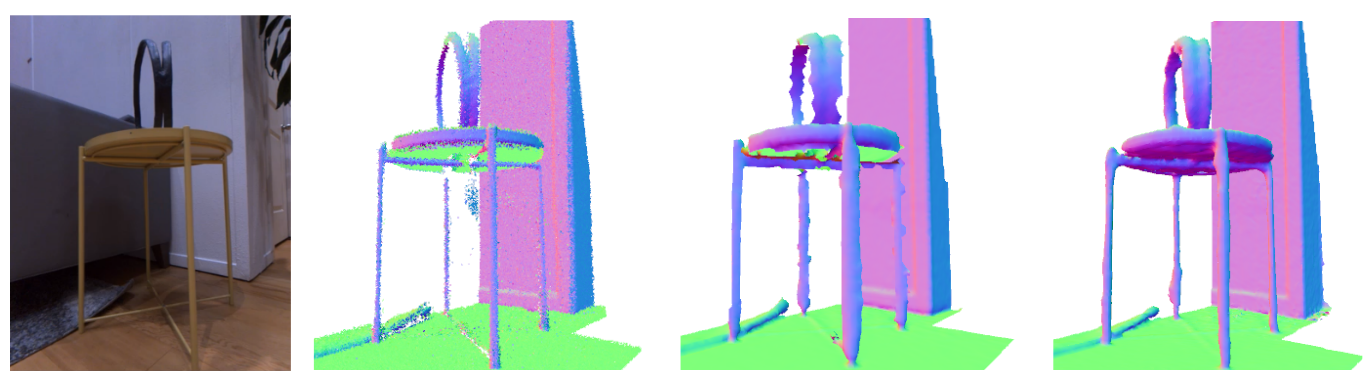}
        %\vspace{-10mm}
        \caption{Close-up view. From left to right: RGB, Input Points, TSR and DeepLS reconstructions.}
        \label{subfig:CloseupStool}
    \end{subfigure}
   
    \caption{\small Qualitative comparison of DeepLS against other 3D Reconstruction techniques on a very challenging thin and incomplete scan. Most of the methods fail to build thin surfaces in this dataset. TSR fits to the thin parts but is unable to complete structures such as the bottom and cylindrical legs of the stool. In contrast, DeepLS reconstructs thin structures and also completes them.}
   \vspace{-5mm}
   
    \label{fig:Stool}
\end{figure}

\vspace{-0.3cm}
Further, we specifically analyse the strength of DeepLS in representing and completing thin local geometry. We collected a scan from an object consisting of two thin circles kept on a stool with long but thin cylindrical legs (see Fig.~\ref{fig:Stool}). The 3D points were generated by a structured light sensor~\cite{whelan2018reconstructing,chabra2019stereodrnet,replica19arxiv}. It was scanned from limited directions leading to very sparse set of points on the stool's surface and legs. We compared our results on this dataset to several 3D methods including TSDF Fusion~\cite{newcombe2011kinectfusion}, Multi-level Partition of Unity (MPU)~\cite{ohtake2005multi}, Smooth Signed Distance Function~\cite{calakli2011ssd}, Poisson Surface Reconstruction~\cite{kazhdan2013screened}, PFS~\cite{ummenhofer2015global} and TSR~\cite{aroudj2017visibility}. We found that due to lack of points and thin surfaces most of the methods failed to either represent details or complete the model. MPU~\cite{ohtake2005multi}, which fits quadratic functions in local grids and is very related to our work, fails in this experiment (see Fig.~\ref{subfig:Stool_MPU}). This indicates that our learned shape priors are more robust than fixed parameterized functions. Methods such as PSR~\cite{kazhdan2013screened}, SSD~\cite{calakli2011ssd} and PFS~\cite{ummenhofer2015global} fit global implicit function to represent shapes. These methods made the thin shapes thicker than they should be. Moreover, they also had issues completely reconstructing the thin rings on top of the stool. TSR~\cite{aroudj2017visibility} was able to fit to the available points but is unable to complete structures such as bottom surface of the stool and it's cylindrical legs, where no observations exist. This shows how our method utilizes local shape priors to complete partially scanned shapes. Please refer to the supplemental materials for further qualitative comparisons and results.

\vspace{-0.1cm}

\vspace{-0.1cm}

\vspace{-2mm}
\section{Conclusion}
\vspace{-2mm}
In this work we presented DeepLS, a method to combine the benefits of volumetric fusion and deep shape priors for 3D surface reconstruction from depth observations. 
A key to the success of this approach is the decomposition of large surfaces into local shapes.
This decomposition allowed us to reconstruct surfaces with higher accuracy and finer detail than traditional SDF fusion techniques, while simultaneously completing unobserved surfaces, all using less memory than storing the full SDF volume would require.
Compared to recent object-centric shape learning approaches, our local shape decomposition leads to greater efficiency for both training and inference while improving surface reconstruction accuracy by an order of magnitude.

\bibliographystyle{splncs04}
\bibliography{egbib}
\newpage
\section*{Supplementary Material}
\setcounter{section}{0}
\renewcommand\thesection{\Alph{section}}
\section{Overview}

The supplementary material consists of detailed information about the experimental setup in Sec.~\ref{sec:experimental_setup}, a detailed view on local shapes in Sec.~\ref{sec:local_shape_interp}, additional metrics for the 3D Warehouse dataset comparison in Sec.~\ref{sec:shapenet_metrics}, and more results, comparisons and details about experiments in Sec.~\ref{sec:real_scene_experiments_app}. Additionally, we provide a video alongside this document, showing reconstructions in motion.

\begin{figure}[h]
     \centering
     \begin{subfigure}{0.62\linewidth}
     \centering
     \includegraphics[width=1\textwidth]{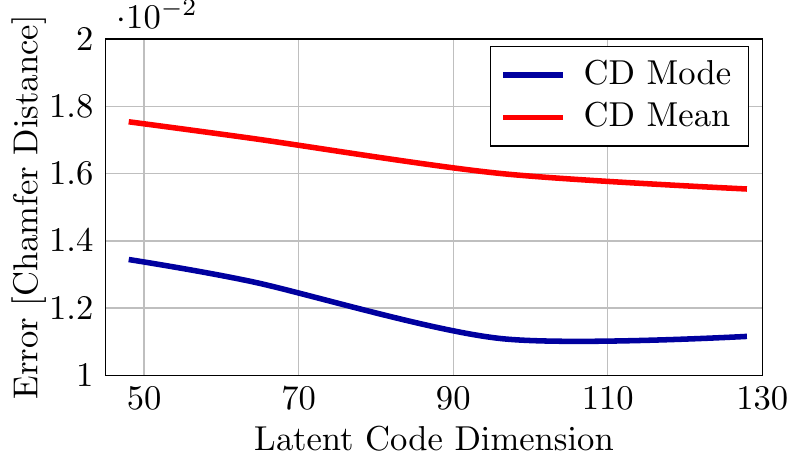}
     \caption{Latent code size}
     \label{fig:CodeParameters}
     \end{subfigure}
     \hfill
     \begin{subfigure}{0.36\linewidth}
     \centering
    \includegraphics[width=1\textwidth]{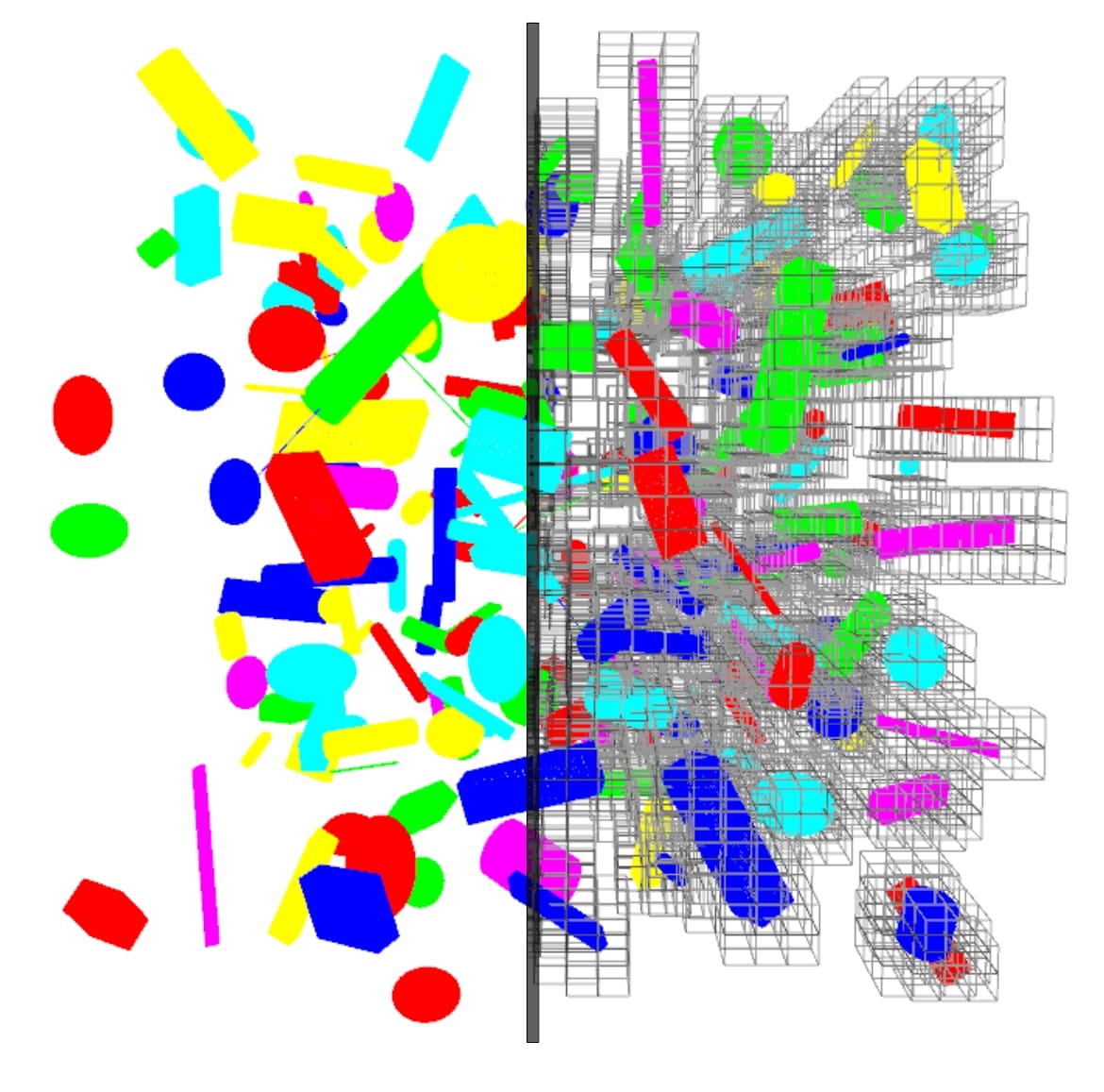}
    \caption{Primitives for training}
    \label{fig:primitives}
    \end{subfigure}
    \caption{Fig. (a) shows the effect of changing the latent code dimensions on the Chamfer distance test error on airplanes class of 3D Warehouse~\cite{Warehouse3D}. Fig. (b) shows an example for a scene containing $200$ primitives shapes as used for training the local shape priors. On the right side, the instantiated local shape blocks are shown.}
\end{figure}

\section{Experimental Setup}
\label{sec:experimental_setup}
\paragraph{\textbf{Autodecoder Network}}
The DeepLS autodecoder network is a lighter version of the network proposed for DeepSDF \cite{park2019deepsdf}. It consists of four fully-connected layers, separated by leaky ReLUs and a tanh at the end, producing values in $[-1,1]$ that are then scaled by the chosen SDF truncation value. Each layer has $128$ output neurons. 
Fig.~\ref{fig:CodeParameters} shows the result of a small study to find the best latent code size in a trade-off between accuracy and compression. We chose a latent size of $125$, leaving us with $128$ input neurons for the first network layer. 

\paragraph{\textbf{Training}}
The output of the network is trained to produce truncated SDF values. To this end, tanh is also applied on the appropriately scaled ground truth SDF before computing the loss against the network output. We chose the scale so that the interval $[-0.9, 0.9]$ after tanh covers approximately two blocks.
We optimize codes and network parameters using the Adam optimizer with initial learning rate of $0.01$, which we decrease twice over the course of training.

\begin{figure}[t]
    \centering
    \includegraphics[width=0.8\linewidth]{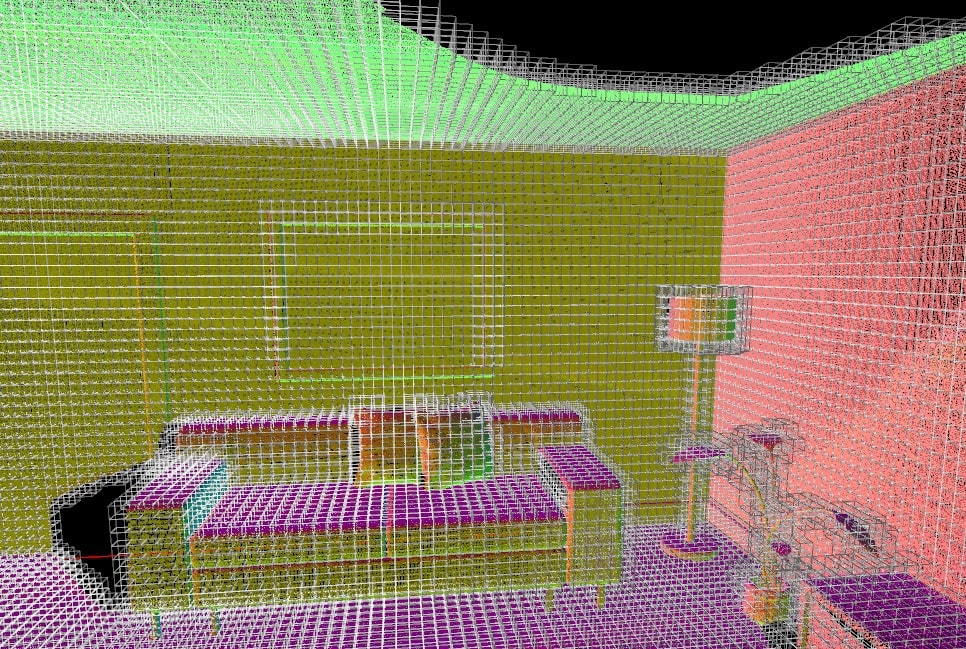}
    \caption{Instantiated local shape blocks in a scene. The blocks are allocated sparsely, based on available depth data, which makes the approach scale well to real world inputs.}
    \label{fig:localblocks}
\end{figure}

\begin{figure*}[t]
    \centering
    \includegraphics[width=1.0\linewidth]{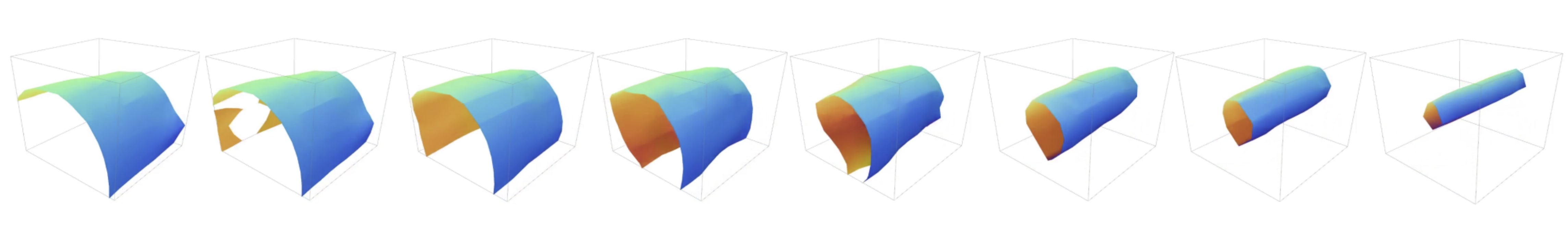}
    \caption{Interpolation in latent space of a local shape code between a flat surface and a pole.}
    \label{fig:latentinterp}
\end{figure*}

\paragraph{\textbf{Training Data}}
The training data to learn local shape priors consists of three different categories of shapes. The first category contains simple primitive shapes, as shown in Fig.~\ref{fig:primitives}, with random 6-DOF pose in space. The second category consists of 3D Warehouse~\cite{Warehouse3D} training meshes: We sampled a subset of $200$ models from each training set of the classes \emph{airplane}, \emph{chair}, \emph{lamp}, \emph{sofa}, and \emph{table}. Each model was split into $32 \times 32 \times 32$ local shape blocks. The last category consists of models from the Stanford 3D scanning repository \cite{StanfordScanRep}, namely \emph{bunny} and \emph{dragon}.

\section{Local Shape Space}
\label{sec:local_shape_interp}
In order to give a better intuition about the space of learned local priors, interpolation sequences between local surfaces are provided in Fig.~\ref{fig:latentinterp}. It should be noted that, in general, the space of possible functions in a voxel is much larger. Therefore, training local priors heavily restricts the space of solutions to those producing local SDF functions that describe reasonable surfaces. The behavior of local shapes over the course of optimization is shown in the accompanying video. Additionally, Fig. \ref{fig:localblocks} show all allocated blocks in a scene, which together reconstruct the whole surface.

\section{Shape border consistency}
\begin{figure}
\centering
\begin{subfigure}{1.0\textwidth}
  \centering
  \includegraphics[width=1.0\textwidth]{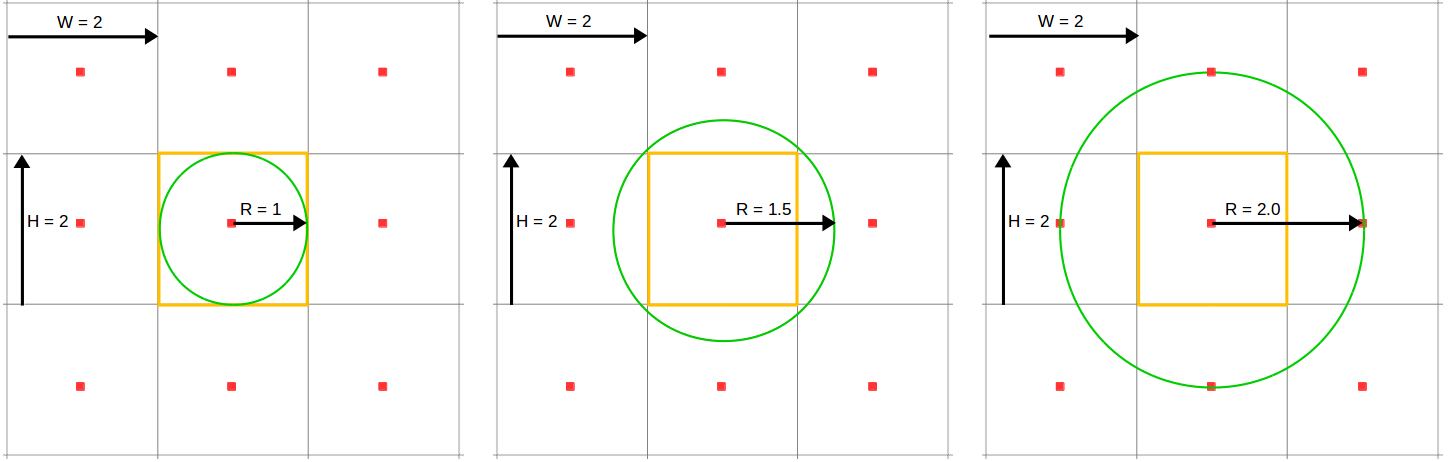}  
  \caption{This figure demonstrates the receptive field of the reference local shape inside yellow block with area inside green circle. $R$ represents the radius of receptive field.}
    \vspace{10pt}
  \label{fig:RFDiag}
\end{subfigure}
\begin{subfigure}{1.0\textwidth}
  \centering
  \includegraphics[width=1.0\textwidth]{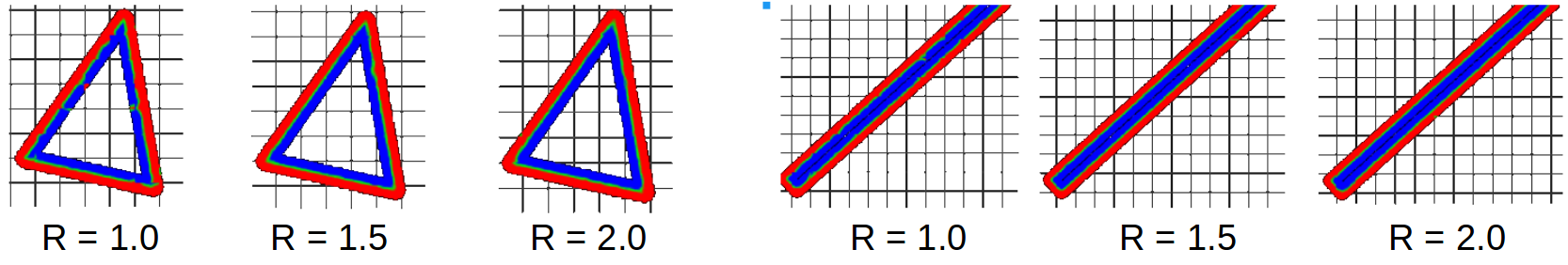}  
  \caption{This figure demonstrates the qualitative difference in the SDF prediction with varying receptive fields.}
  \label{fig:RFResults}
\end{subfigure}
\begin{subfigure}{0.45\textwidth}
  \centering
  \includegraphics[width=1.0\linewidth]{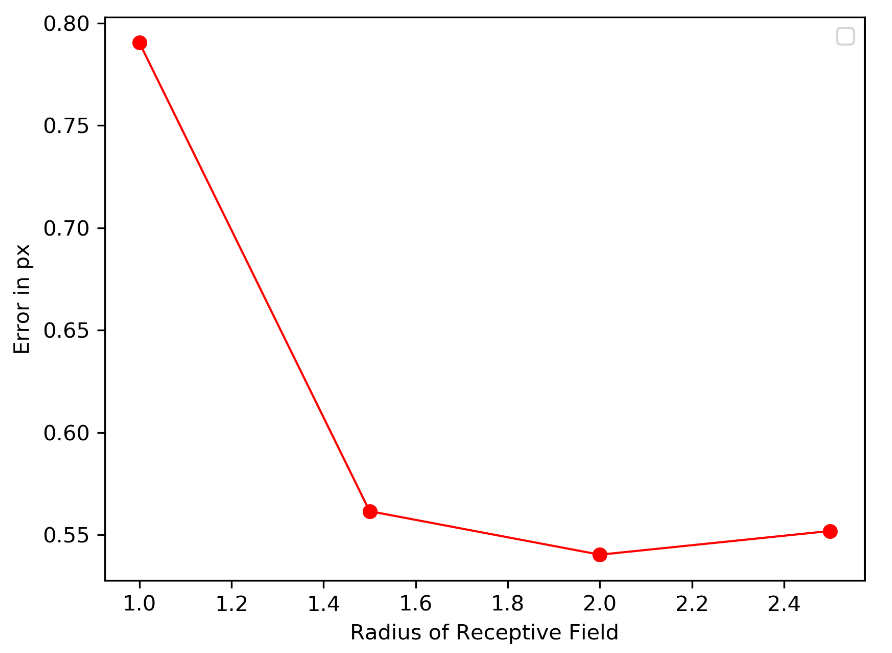}  
  \caption{Error in SDF prediction with increasing receptive field. Critical point in receptive field is observed.}
  \label{fig:RFPlot}
\end{subfigure}
\qquad
\begin{subfigure}{0.45\textwidth}
  \centering
  \includegraphics[width=1.0\linewidth]{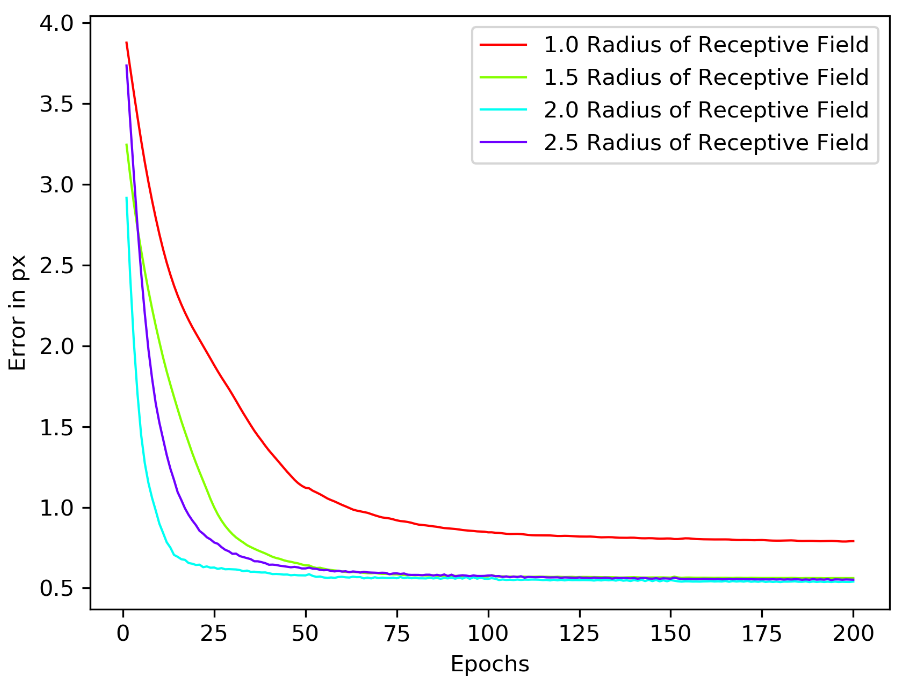}  
  \caption{Plot shows the test error with increasing iterations in optimization. Optimal receptive field shows early convergence.}
  \label{fig:RFPlotOptimization}
\end{subfigure}
\caption{This figure demonstrates the effect of receptive field on the quality of reconstruction of SDF. }
\label{fig:RF}
\end{figure}

In order to better understand the border consistency among the borders of local shapes we used simple 2D scenes often composed of simple primitive shapes such as triangles, rectangle and circles. In training and testing session we sample points around these shapes and extract SDF measurements as described in DeepSDF~\cite{park2019deepsdf}. Note, we color code these sample points with red for positive, blue for negative and green for zero SDF measurements. In all the 2D experiments we use roughly 1000 samples inside a grid cell (local shape spatial size) in training session and 100 samples in test session. We report testing error as the SDF prediction error in 2d (pixels).\\
In the following experiment, in order to study shape border consistency we increased the receptive field of local shapes as shown in Figure~\ref{fig:RFDiag}. By receptive field we mean the physical space of input samples for a particular local shapes. In general, we observe improvement in SDF prediction on the boundaries of local shapes with increasing receptive field as shown in Figure~\ref{fig:RFResults}. Although, we observe a critical point in the receptive field after which the performance drops as shown in Figure~\ref{fig:RFPlot}. As increasing receptive field makes the local shapes bigger and more complicated so more parameters in the network $F_{\theta}$ are required to express the space of shapes. Hence, each $F_{\theta}$ has a critical point in the receptive field. We also observe the early convergence in optimization for optimal receptive field as shown in Figure~\ref{fig:RFPlotOptimization}.

\section{3D Warehouse Comparison - Additional Metric}
\label{sec:shapenet_metrics}
In Tab~\ref{tab:shapenetTest_app} we extend the comparison on shapes from 3D Warehouse dataset on other metrics. In addition to the Chamfer distance we show mesh accuracy, which is defined as the maximum distance $d$ such that $90\%$ of generated points are within $d$ if the ground truth mesh. All metrics show the similar trend that DeepLS achieves way higher accuracy than the related object-level representations.
 \begin{table}
\footnotesize
\centering
 \begin{tabular}{l|r|r|r|r|r}
 \toprule
 %\hhline{|=|=|=|=|=|=|}
  \multicolumn{1}{l|}{CD, mean} & \multicolumn{1}{c|}{chair} & \multicolumn{1}{c|}{plane} & \multicolumn{1}{c|}{table} & \multicolumn{1}{c|}{lamp} & \multicolumn{1}{c}{sofa}  \\
 \midrule
 AtlasNet-Sph.  & 0.752 & 0.188 & 0.725 & 2.381 & 0.445\\
 AtlasNet-25      & 0.368 & 0.216 & 0.328 & 1.182 & 0.411\\
 DeepSDF        &  0.204 & 0.143 & 0.553 & 0.832 & 0.132  \\
  DeepLS        &  \bf{0.030} & \bf{0.018} & \bf{0.032} & \bf{0.078} & \bf{0.044}  \\
 \midrule
  \multicolumn{1}{l}{CD, median} &  \multicolumn{1}{c}{ \ } &  \multicolumn{1}{c}{ \ } &  \multicolumn{1}{c}{ \ } &  \multicolumn{1}{c}{ \ }  &  \multicolumn{1}{c}{} \\
 \midrule
  AtlasNet-Sph.  & 0.511 & 0.079 & 0.389 & 2.180 & 0.330 \\
 AtlasNet-25      & 0.276 & 0.065 & 0.195 & 0.993 & 0.311  \\
 DeepSDF        &  0.072 & 0.036 & 0.068 & 0.219 & 0.088  \\
 DeepLS        & \bf{0.023} & \bf{0.011} & \bf{0.026} & \bf{0.019} & \bf{0.039}  \\
 \midrule
%  \multicolumn{1}{|l}{EMD, mean} &  \multicolumn{1}{c}{ \ } &  \multicolumn{1}{c}{ \ } &  \multicolumn{1}{c}{ \ } &  \multicolumn{1}{c}{ \ }  &  \multicolumn{1}{c|}{} \\
% \hline
%  AtlasNet-Sph.  & 0.071 & 0.038 & 0.060 & 0.085 & 0.050  \\
% AtlasNet-25      & 0.064 & 0.041 & 0.073 & 0.062 & 0.063  \\
% DeepSDF        &  \bf{0.049} & \bf{0.033} & \bf{0.050} & \bf{0.059} & \bf{0.047}  \\
% \hline

  \multicolumn{1}{l}{Mesh acc., mean} &  \multicolumn{1}{c}{ \ } &  \multicolumn{1}{c}{ \ } &  \multicolumn{1}{c}{ \ } &  \multicolumn{1}{c}{ \ }  &  \multicolumn{1}{c}{} \\
 \midrule
  AtlasNet-Sph.  & 0.0330 & 0.0130 & 0.0320 & 0.0540 & 0.0170  \\
 AtlasNet-25      & 0.0180 & 0.0130 & 0.0140 & 0.0420 & 0.0170  \\
 DeepSDF        & 0.0090 & 0.0040 & 0.0120 & 0.0130 & 0.0040  \\
 DeepLS        & \,\,\,\,\bf{0.0009} & \,\,\,\,\bf{0.0008} &\,\, \,\,\bf{0.001} & \,\,\,\,\bf{0.0012} & \,\,\,\,\bf{0.0011} \\
 \bottomrule

   \end{tabular}
   \caption{\small Representing unknown shapes from the 3D Warehouse~\cite{Warehouse3D} test set. In addition to the Chamfer distance, we provide mesh accuracy~\cite{seitz2006comparison}.  Lower is better for all metrics. It can be seen that all metrics show a similar trend.} 
   \label{tab:shapenetTest_app}
\end{table}

\begin{figure}[t]
    \centering 
    \begin{subfigure}[t]{0.49\linewidth}
        \centering 
        \includegraphics[width=\textwidth,clip,trim=0 100 0 150]{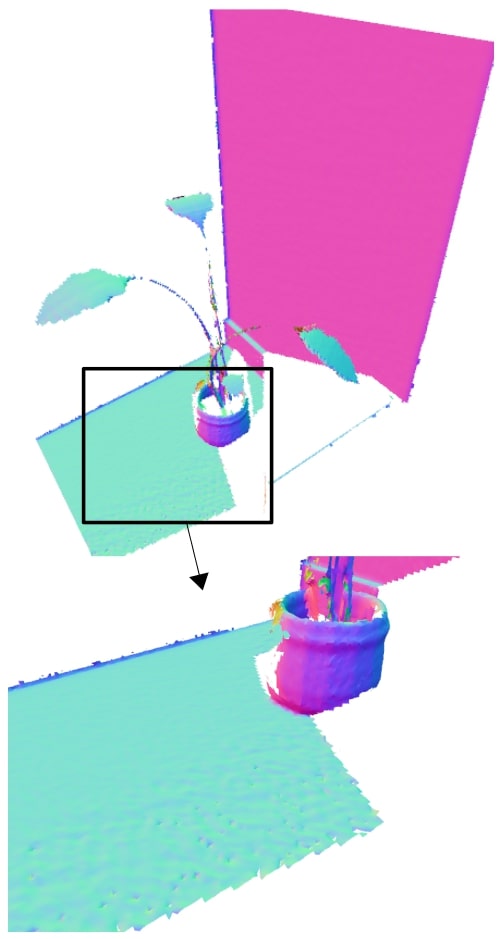}
        \caption{TSDF Fusion} 
    \end{subfigure}
    \hfill
    \begin{subfigure}[t]{0.49\linewidth}
        \centering 
        \includegraphics[width=\textwidth,clip,trim=0 100 0 150]{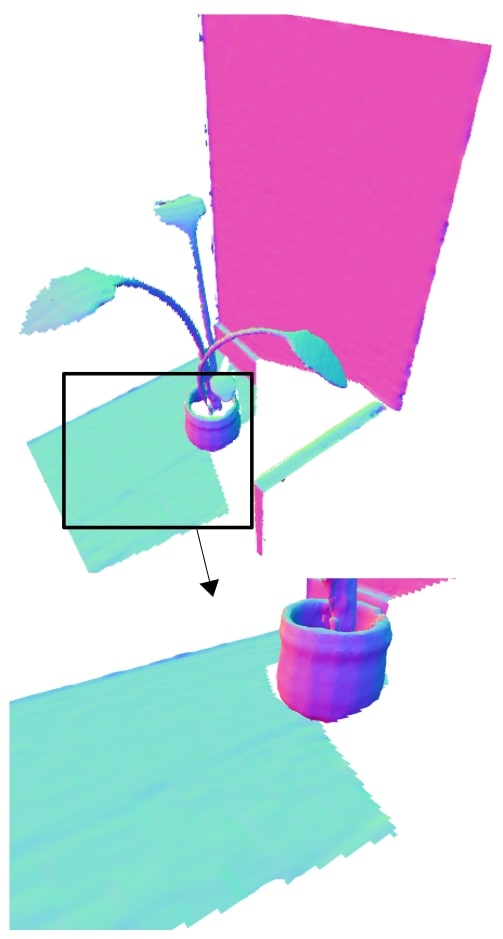}
        \caption{DeepLS}
    \end{subfigure}
   
    \caption{\small The figure shows a part of the ICL-NUIM kt0 scene~\cite{handa:etal:ICRA2014}, reconstructed from samples with artitificial noise of $\sigma = 0.015$. DeepLS shows better denoising properties than TSDF Fusion. For the whole ICL-NUIM benchmark scene, DeepLS achieves a surface error of \textbf{6.41} mm with \textbf{71.04} \% completion while TSDF Fusion has an error of 7.29 mm and 68.53 \% completion.}
    \label{fig:ICLNUIMnoise}
\end{figure}

\section{Scene Experiments}
\label{sec:real_scene_experiments_app}
Here, we explain the process from depth maps to SDF samples for real scenes in more detail and provide qualitative results. See also the provided video for further results.

\begin{figure}
    \centering
    \begin{subfigure}[t]{\linewidth}
    \includegraphics[width=\linewidth]{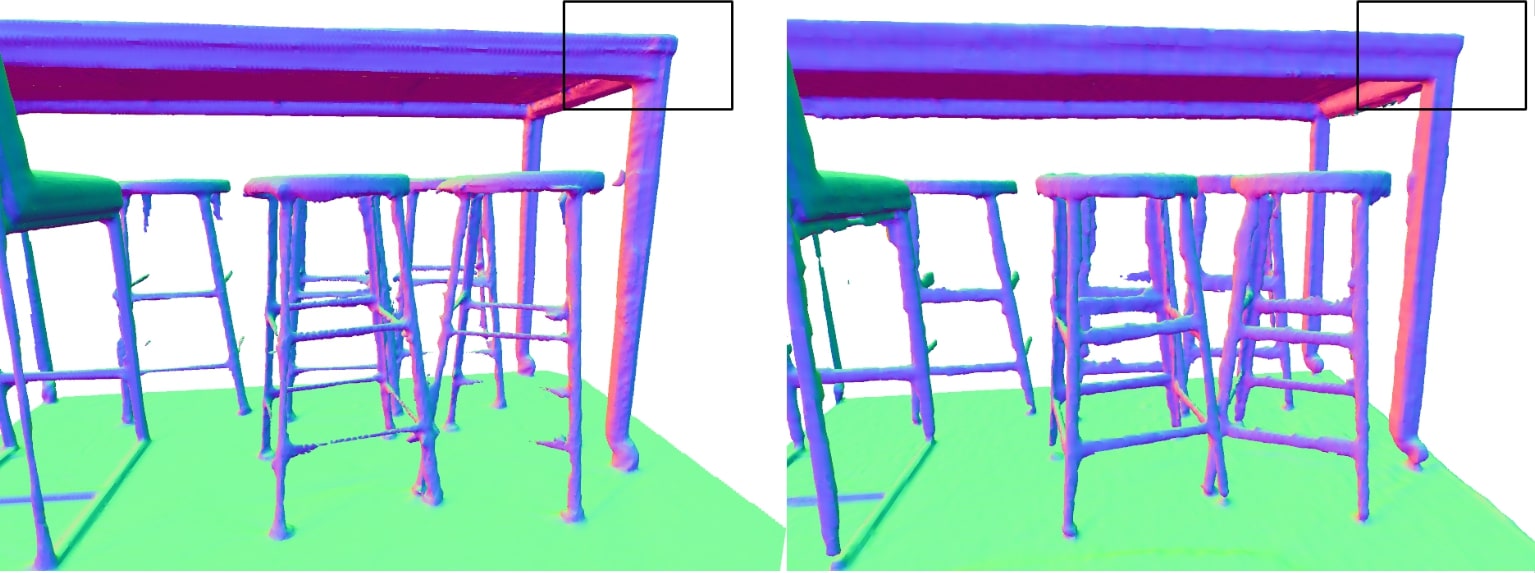}
    
    \includegraphics[width=\linewidth]{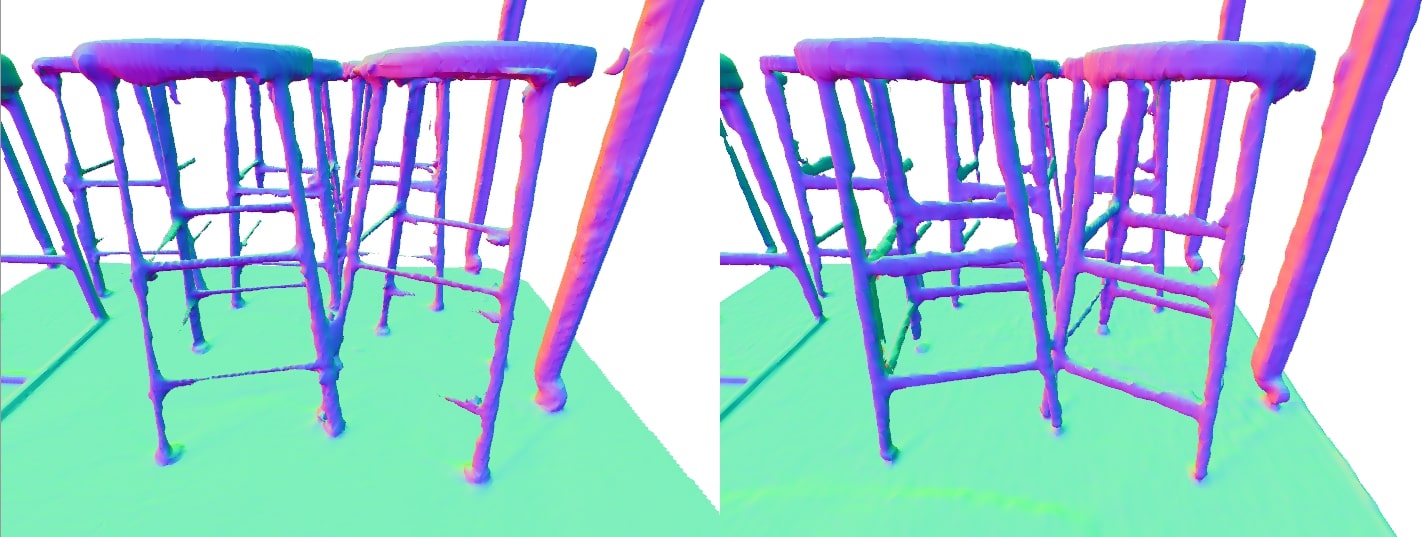}
    \caption{DeepLS (right) captures thin chair legs better than TSDF Fusion (left) which tends to loose those details.}
    \end{subfigure}

    \begin{subfigure}[t]{\linewidth}
    \includegraphics[width=\linewidth]{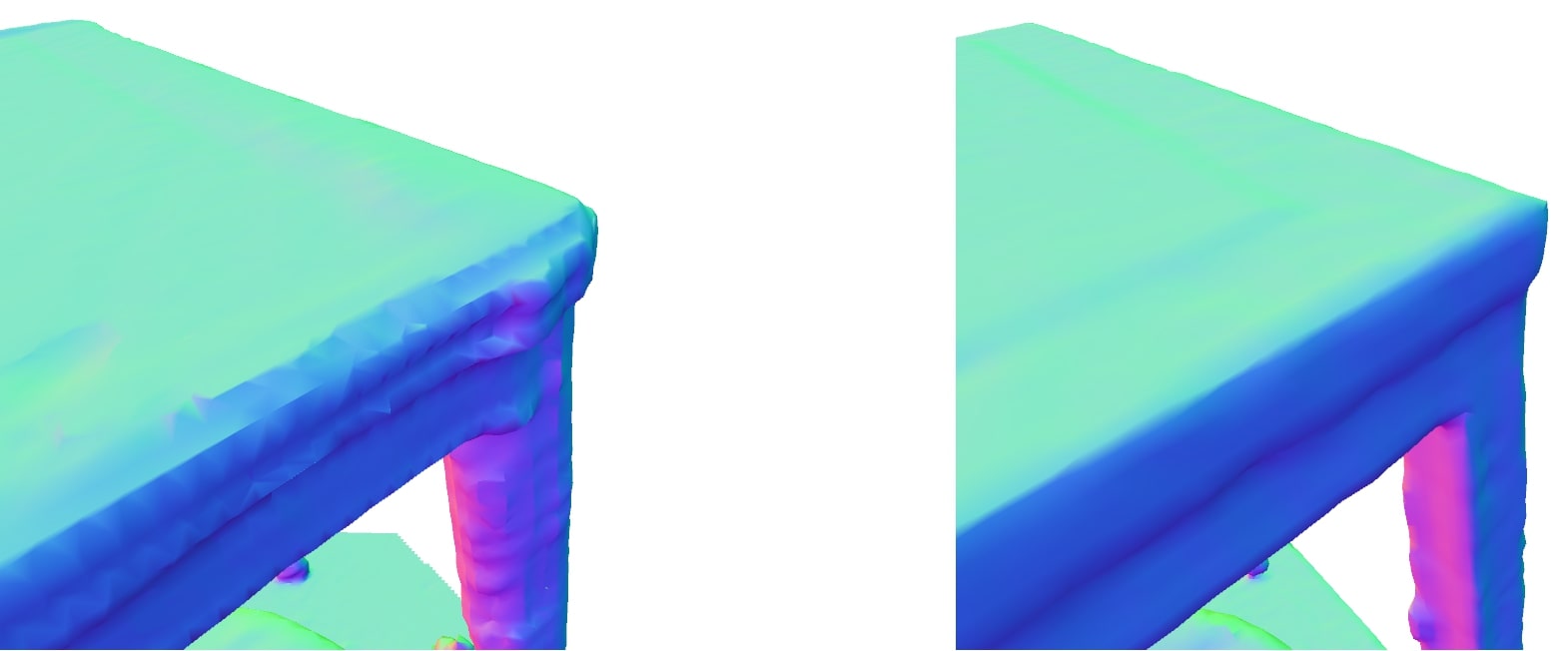}

        \caption{Zoomed view of region marked with black box in (a). DeepLS (right) represents sharper corners and smoother planes than TSDF Fusion (left).}
        \label{fig:realSceneChairsCorner}
    \end{subfigure}
    \caption{Qualitative comparison of TSDF Fusion (left) with DeepLS (right) on real scanned data prepared using a structured light sensor system~\cite{replica19arxiv}. The figure (b) is the magnified region marked with black box in figure (a).}
    \label{fig:realSceneChairs}
\end{figure}

\subsection{Sample Generation}
Sample generation from depth scans consists of the following steps: (1) For a given scene, we generate a collection of 3D points from depth maps. (2) For each depth point, we create one sample with zero SDF, and several positive and negative SDF samples by moving the sample along the pre-computed surface normal by $1.5$ cm and $-1.5$ cm, respectively. The accompanying SDF value is chosen as the moved distance. 

(3) We generate additional free space samples along the observation rays. Further, we weight each set of points inversely based on the depth of the initial scan point, to ensure that accurate points closer to the scanning device are weighted higher. This procedure is described in detail in TSDF Fuison~\cite{curless1996volumetric}.
Similar to traditional SDF fusion approaches~\cite{newcombe2011kinectfusion}, DeepLS exposes a parameter which controls the size of the region around actual depth samples in which marching cubes is performed. Varying this parameter leads to the mesh accuracy vs. completion trade-off, discussed in the main paper.

\begin{figure}
    \begin{subfigure}[t]{0.49\linewidth}
        \centering 
        \includegraphics[width=\textwidth]{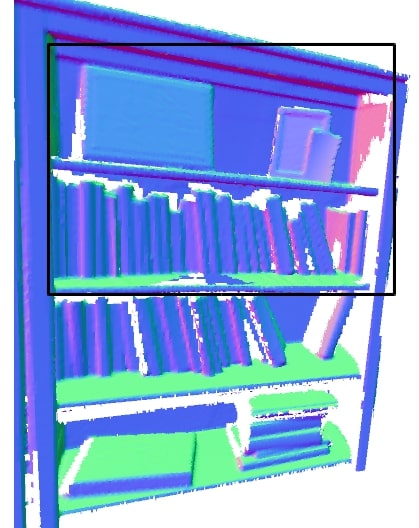}
    \end{subfigure}
   \begin{subfigure}[t]{0.49\linewidth}
        \centering 
        \includegraphics[width=\textwidth]{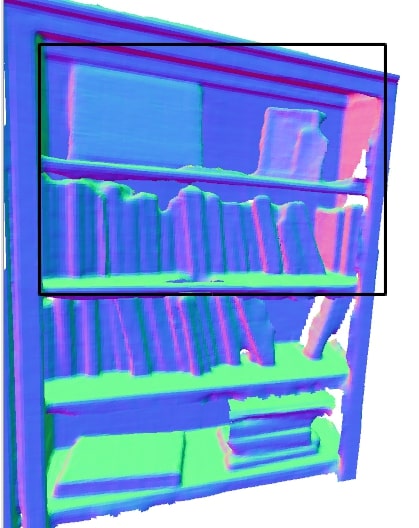}
    \end{subfigure}
    
     \begin{subfigure}[t]{0.49\linewidth}
        \centering 
        \includegraphics[width=\textwidth]{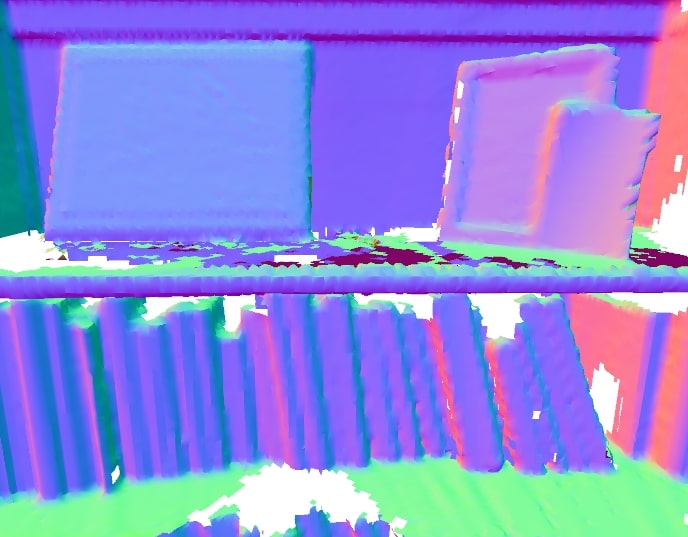}
        \caption{TSDF Fusion~\cite{curless1996volumetric}} 
    \end{subfigure}
   \begin{subfigure}[t]{0.49\linewidth}
        \centering 
        \includegraphics[width=\textwidth]{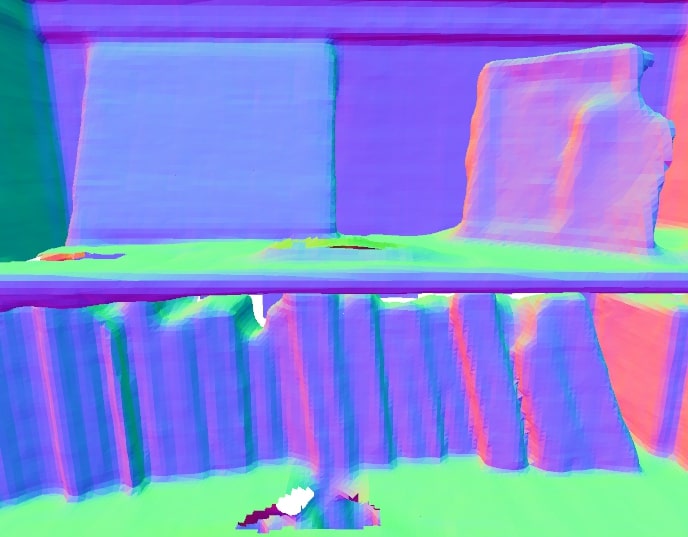}
        \caption{DeepLS} 
        \label{fig:RealSceneDeepLS} 
    \end{subfigure}

    \caption{\small We show the scene reconstruction quality of DeepLS vs TSDF Fusion~\cite{curless1996volumetric} on a partially scanned real scene dataset using a structured light sensor system~\cite{replica19arxiv}. This figure shows that DeepLS provides better local shape completion than TSDF Fusion. The bottom row represents the zoomed in view marked with black box in the top row.}
    \label{fig:RealScene} 
\end{figure}

\subsection{Comparisons for Synthetic Noise}
Fig.~\ref{fig:ICLNUIMnoise} shows results of DeepLS and TSDF Fusion on an ICL-NUIM benchmark scene with artificial noise of $\sigma = 0.015$. The learned local shape priors of DeepLS effectively are able to find plausible surfaces given the noisy observations, which results in smoother surfaces in comparison to TSDF Fusion.

\subsection{Qualitative Results}
We show additional qualitative results on real scanned data in Fig.~\ref{fig:realSceneChairs}, Fig.~\ref{fig:RealScene} and in the supplemented video. Both scenes showed in the figures were captured using a handheld structured light sensor system as was used for capturing the Replica dataset~\cite{replica19arxiv} and in related work~\cite{whelan2018reconstructing,chabra2019stereodrnet}. An in-house SLAM system, similar to state-of-the-art systems~\cite{engel2017direct,mur2015orb}, was used to provide 6 degree of freedom (DoF) poses for individual depth frames from the sensor. It can be seen that DeepLS succeeds in representing small details like the bars of chairs while TSDF Fusion tends to loose these details. Also, we observe sharper corners (c.f.~\ref{fig:realSceneChairsCorner}) and more complete surfaces (c.f.~\ref{fig:RealSceneDeepLS}) with DeepLS.

\end{document}